\documentclass[11pt]{article}

\usepackage[preprint]{acl}

\usepackage{kurier}

\usepackage{times}
\usepackage{latexsym}

\usepackage[T1]{fontenc}
\usepackage[utf8]{inputenc}
\usepackage{xcolor}

\usepackage{microtype}

\usepackage{inconsolata}

\usepackage{graphicx}

\usepackage{xspace}
\usepackage{xcolor}
\usepackage[table]{xcolor}
\usepackage{enumitem}
\usepackage{booktabs}
\usepackage{makecell}
\usepackage{multirow}
\usepackage{subcaption}
\usepackage{mdframed}
\usepackage{tabularx}
\usepackage{amssymb}
\usepackage{ulem}
\usepackage{fontawesome5}

\definecolor{overallpref}{HTML}{1896DE}
\definecolor{nearexp}{HTML}{F27200}
\definecolor{deepexp}{HTML}{006C65}
\definecolor{lightgray}{gray}{0.9}
\newcolumntype{L}[1]{>{\raggedright\let\newline\\\arraybackslash\hspace{0pt}}p{#1}}
\newcolumntype{M}[1]{>{\raggedright\let\newline\\\arraybackslash\hspace{0pt}}m{#1}}

\definecolor{darkgreen}{rgb}{0.0, 0.4, 0.13}
\definecolor{lightlightblue}{rgb}{0.9, 0.95, 1.0}
\definecolor{mustard}{rgb}{0.9, .61, .11}

\newcommand{\metricwise}{{\fontsize{10}{8} \fontfamily{kurier}\selectfont \textit{metric-wise}} \xspace}
\newcommand{\metricwisecap}{{\fontfamily{kurier}\selectfont \textit{Metric-wise}}\xspace}
\newcommand{\preferencerank}[1]{{\fontsize{10}{8} \fontfamily{kurier}\selectfont \textit{{#1}}}\xspace}

\newcommand{\sqaeval}{\textcolor{darkgray}{\texttt{ScholarQA-CS2}}\xspace}
\newcommand{\sqaevalbl}{{\texttt{ScholarQA-CS2}}\xspace}
\newcommand{\astabench}{AstaBench\xspace}

\title{Deep Research, Shallow Evaluation: \\A Case Study in Meta-Evaluation for Long-Form QA Benchmarks}

\author{{Jena D. Hwang \quad 
        Varsha Kishore \quad
        Amanpreet Singh \quad
        Dany Haddad %\enskip
        }\\
        {\bf
        Aakanksha Naik \quad
        Malachi Hamada \quad      
        Jonathan Bragg \quad
        Mike D'Arcy %\enskip
        }\\        
        {\bf
        Daniel S. Weld \quad
        Lucy Lu Wang \quad
        Doug Downey \quad
        Sergey Feldman
        }        
        \\ [1mm]
        Allen Institute for AI
        \\
        {\tt\small \{jenah, sergey\}@allenai.org}
}

\begin{document}
\maketitle
\begin{abstract}

%In the recent year,  systems generating long-form reports have become widely available. Correspondingly, various evaluation frameworks have emerged to assess these systems, converging on methods like LLM-as-judge protocols and claim verification.
Recent advances have made long-form report–generating systems widely available. 
This has prompted evaluation frameworks that use LLM-as-judge protocols and claim verification, along with \textbf{meta-evaluation} frameworks that seek to validate these methods.
Many of the meta-evaluations estimate an evaluation quality's by comparing its assessments against human pairwise preferences.
Prior work, however, suggests that human pairwise preference may be overly simplistic and can fail to capture nuances of expert expectations. 
We conduct a case study in meta-evaluation for long-form QA benchmarks using \sqaeval,%\footnote{Benchmark and its citations are anonymized for review; \sqaeval and \astabench are anonymized names. \label{footnote:anonymization}} 
a benchmark designed for assessing retrieval-augmented deep-research QA in the scientific domain.
We comprehensively validate the benchmark through human pairwise preference judgments, then critically examine the strengths, weaknesses, and confounders of this approach.  We show that pairwise preference rankings are best suited for system-level evaluation, while explicit metric-wise annotations and expert annotators are critical for reliable metric-level assessment, with subjectivity remaining a key challenge. Based on our findings, we offer practical guidelines for designing future meta-evaluations that better align evaluation methods, annotator expertise, and reporting practices. By surfacing these methodological challenges, we aim to advance evaluation standards for deep-research systems. 

%However, prior work and our own evaluation of these benchmarks indicate that they do not consistently align with expert human annotations. 

%maybe rewrite to: "However, prior work and our own evaluation of these benchmarks indicate that they do not consistently align with expert human annotations. We conduct a meta-evaluation of our recently-released benchmark, \sqaeval, designed for assessing retrieval-augmented deep-research QA in the scientific domain. Through multiple meta-evaluation designs comparing LLM judgments against human preferences, we critically examine the strengths, weaknesses, and confounders of this approach..."  

\end{abstract}

\section{Introduction}

{\let\thefootnote\relax\footnotetext{\faGithub~\href{https://github.com/allenai/ai2-scholarqa-eval/}{\footnotesize \texttt{allenai/ai2-scholarqa-eval}}}}
In the last few years, we have witnessed rapid progress of deep-research systems, with the release of many automated systems for long-form QA. Both proprietary and open services such as OpenAI Deep Research \cite{openai2025deepresearch}, 
%leaving this out for anonym version. because it is conspicuously different from others and adding storm as the open one in the intro.
%ScholarQA \cite{singh2025ai2scholarqaorganized}, 
Elicit \cite{elicit2024}, Perplexity \cite{perplexity2024}, ScholarQA \cite{singh2025ai2scholarqaorganized}, and
Storm \cite{shao2024assistingwritingwikipedialikearticles} provide agentic implementations that plan, retrieve, and synthesize a large collection of documents to produce long-form reports for user queries.
%\dd{cite these systems (can borrow bibtex from Dany's paper)}. 
In response, a growing collection of evaluation frameworks and benchmarks have emerged to assess the quality of system generations, introducing methods such as LLM-as-a-judge protocols, rubric-based metrics, and claim verification.
With many of the recent frameworks relying on large language models (LLMs) as evaluation judges, it has become standard practice to validate these automated judgments through comparisons with human pairwise preferences~\citep{Bosse2025deepresearchbench,yifei2025researchqa}, which we call \textit{overall preference ranking}.
% \vk{can we explicitly state that that this is overall prefernce?}. 
These validation efforts rest on the 
%implicit 
assumption that when a benchmark can precisely enumerate aspects of a ``good'' report and measure them successfully, these measures should meaningfully approximate human expectations for quality reports. 

% narrow the scop to specifcally pick on human PREFERENCE-based evaluation.
The reality, however, is that these contemporary deep-research benchmark validation practices tend to be overly simplistic in nature, treating alignment with overall human preference ranking as sufficient despite existing evidence that human preference is multi-faceted and context-dependent \citep{hu-etal-2023-decipherpref, Warren2011ValuesAP}  and that such alignment can fail to capture the nuances of expert expectations \citep{Zhao2025SciArenaAO,Szymanski2024LimitationsOT, wang-etal-2023-automated,  Xu2023ACE}.
%its apparent inability to reflect the nuances of expert expectations 
%\dd{do these citations correspond to the overly simple practices, or to the documented inability of those to reflect nuances?  Perhaps rephrase to clarify}. 
An underlying issue is that we lack systematic understanding of the factors that drive this misalignment: how different forms of human annotation protocols, varying levels of annotator expertise, or other inherent subjective factors shape the apparent agreements between humans and LLM-judged  metrics. As a result, we are left to draw meaningful inferences from numeric values with `higher-is-better' expectations, while lacking clear understanding of whether strong system-level agreement with human preference truly indicates that the employed metrics capture the intended dimensions of quality. 

In this work, we address this gap by investigating utility and pitfalls of using pairwise human preference judgment for \textbf{meta-evaluation} of long-form research benchmarks within the scientific domain. We ask: to what extent do current LLM-based evaluation metrics really reflect expert judgments about different dimensions of answer quality, and how does this depend on (i) how we collect those human judgments and (ii) how versed the experts are in the specific focus of the questions?

We focus our meta-evaluation efforts on \sqaeval, a recent benchmark that evaluates deep research agents based on user-input research questions and system-generated long-form responses. We compare the benchmark's results against human preference rankings to surface the merits and shortcomings of the meta-evaluation method. We also carefully design additional expert annotation settings to investigate effects of metric-wise annotations and depth of annotator expertise on evaluation results, taking stock of potential confounders and surfacing key caveats in these design choices. Through this study, we aim to identify limitations and challenges of our current meta-evaluation approaches, and provide recommendations for improving evaluation practices in the field. Our contributions are as follows:
%We summarize our main contributions below:

\begin{itemize}[itemsep=0pt, leftmargin=.2in, topsep=0pt, parsep=1pt]
\item We present comprehensive meta-evaluation of the recent deep-research benchmark, \sqaeval. Our present study is the first to examine how deep-research evaluation accuracy can vary when assessing overall system performance compared to individual instance or metric performance, and how annotator expertise influences evaluation accuracy and perceived subjectivity.
\item Our analysis reveals important findings, including: (1) comparisons to human pairwise evaluation is best suited for system-level evaluation, (2) explicit metric-wise annotation is necessary for fine-grained assessment, (3) depth of annotation expertise impacts their assessment, and (4) human evaluators exhibit notable subjectivity that underscore the challenge of the task. We also show these results hold consistently across different LLM judges. 
\item Based on our findings, we provide practical guidance for future meta-evaluations of deep research benchmarks that emphasize closer attention to metrics assessed, annotator expertise, and reporting practices. 
%emphasizes human pairwise preferences for system evaluation, task-specific annotations for metric assessment, and reporting instance level agreements alongside aggregate scores.
%We recommend human pairwise preference judgments be used for system-level evaluation, while relying on task-specific human annotations that mirror LLM evaluators for metric-level assessment, emphasize the importance of aligning annotator expertise with evaluation goal, and encourage reporting disagreements alongside aggregate scores to account for subjectivity and user context in evaluating complex research outputs.
\end{itemize}

\begin{table*}[th!]

\renewcommand{\arraystretch}{1.15}
\setlength{\tabcolsep}{3pt}
\footnotesize
%p{10mm}p{6mm}p{10mm}p{11mm}
\rowcolors{3}{}{lightgray!50}
\begin{tabularx}{\textwidth}{@{}p{29mm}lp{12mm}p{5mm}p{9mm}ccccc|ccc@{}}
\rowcolor{white}       &  &  & & &  \multicolumn{8}{c}{\textbf{Human Meta-Evaluation}}  \\ 
\rowcolor{white}       &  &  & & &  \multicolumn{8}{c}{\textbf{(Expert Agreement w/ LLM-judge)}}  \\ \cmidrule{6-13}
\rowcolor{white} 
{\fontsize{8}{8}\makecell[lt]{\\\textbf{Evaluation}}}  &   
{\fontsize{8}{8}\makecell[lt]{\\\textbf{Dom}}}      & 
{\fontsize{8}{8} \makecell[lt]{\textbf{Input}\\\textbf{Source}}} & 
\textbf{Ret Src} & 
{\fontsize{8}{8}\makecell[lt]{\textbf{Metrics}\\\textbf{Used}}} & 
{\fontsize{8}{8} \makecell[lt]{\textbf{Expert}\\\textbf{Eval}}}      &  
{\fontsize{8}{8} \makecell[lt]{\\\textbf{PPR}}} & 
{\fontsize{8}{8} \makecell[lt]{\textbf{Metric}\\\textbf{-Wise}}} & 
{\fontsize{8}{8} \makecell[lt]{\textbf{Reported}\\\textbf{Comp.}}} & 
{\fontsize{8}{8} \makecell[lt]{\textbf{Qs}\\\textbf{Assign}}}   &
{\fontsize{8}{8} \makecell[lt]{\textbf{\# Sys}\\\textbf{Comp.}}} & 
{\fontsize{8}{8} \makecell[lt]{\\\textbf{IAA}}} &
{\fontsize{8}{8} \makecell[lt]{\\\textbf{OA}}}\\ \midrule

\rowcolor{yellow!20} \makecell[tl]{\sqaevalbl\\ \citep{bragg2025astabenchrigorousbenchmarkingai} \\ \ + present study}   & Sci:CS            & \makecell[tl]{real user\\queries}                     & \texttt{\small PI}                               & \texttt{\small AR,RCC, CP,CR }       &   $\checkmark$   &  $\checkmark$ &  $\checkmark$$\checkmark$ & \makecell[tc]{\texttt{~SC,IC,~~}\\\texttt{~OA,MA,~~}\\\texttt{MC}}& \texttt{\small R,C,A}  & 6 & 55.0\% & 51.6\%\\
\makecell[tl]{OpenScholar\\ \citep{asai2024openscholar}}                  & Sci          & written queries              & \texttt{\small PI}                                               & \texttt{\small RCC,CP, CR,AQ}        &  $\checkmark$  & $\checkmark$  & $\checkmark$* & \texttt{\small OA} & \texttt{\small R}  & 3 & 68\% & 50\%\\
\makecell[tl]{DeepResearchGym\\\citep{coelho2025deepresearchgym}}               & Gen & \makecell[tl]{real user\\queries}  & \texttt{\small WI}                  & \texttt{\small FC,AQ, CP,CR }          & $\checkmark$ & $\checkmark$ & $\checkmark$* & \texttt{\small SC,OA} & \texttt{\small R}  & 7 & \makecell[tc]{0.87\\{\scriptsize Cohen $\kappa$}}  & \makecell[tc]{0.72\\{\scriptsize Cohen $\kappa$}}\\
\makecell[tl]{DeepResearch Bench\\\citep{du2025deepresearch}}                   & Gen             & curated queries       & \texttt{\small WS}                                                 & \texttt{\small RC,CP, CR}          &  $\checkmark$  & $\checkmark$ &  - & \texttt{\small SC,OA} & \texttt{\small R} & 4 & 68.4\% & 71.3\% \\
\makecell[tl]{\scriptsize DeepResearch-ReportEval\\\citep{Fan2025UnderstandingDV}} & Gen  &  curated queries & \texttt{\small GT, WS}      & \texttt{\small FC,AQ}           &  $\checkmark$  & $\checkmark$  & - & \texttt{\small OA} & \texttt{\small R} & 4 & - & 61.1\%\\
\makecell[tl]{ResearchQA       \\\citep{yifei2025researchqa}}                 & Gen           & synthetic queries      & \texttt{\small WS}                                                  & \texttt{\small RCC,CR}                &   $\checkmark$ & $\checkmark$ &  $\checkmark$ & \texttt{\small OA,MA}  & \texttt{\small R}  & 3 & 84\% & 75\% \\
\makecell[tl]{\scriptsize DeepResearch Bench II\\\citep{Li2026DeepResearchBI}}                   & Gen             & \makecell[tl]{curated\\tasks}         & \texttt{\small WS}                                                 & \texttt{\small RC,CP, CR}          &  $\checkmark$  & -  &  $\checkmark$* & \texttt{\small OA} & \texttt{\small R}  &&&\\
\makecell[tl]{DeepScholar-Bench\\\citep{patel2025deepscholarbench}}               & Gen          & \makecell[tl]{extracted\\ abstracts}                & \texttt{\small GT, WS}                                     & \texttt{\small FC,CP, CR}              &   $\checkmark$ & - &  $\checkmark$* & \texttt{\small OA} & \texttt{\small R}  &&&\\
\makecell[tl]{Trec Rag 2024\\\citep{pradeep2024trecrag}}  & Gen                   & \makecell[tl]{written\\ topics}               & \texttt{\small GT}                                                     & \texttt{\small FC}                      &               $\checkmark$ & - & $\checkmark$ & \texttt{\small SC,MC} & \texttt{\small R}  &&&\\
\makecell[tl]{DeepResearchEval   \\\citep{Wang2026DeepResearchEvalAA}}                  & Gen           & \makecell[tl]{synthetic \\tasks}     & \texttt{\small WS}                 & \texttt{\small FC,RC, CP}              & $\checkmark$ & - & $\checkmark*$ & \texttt{OA} & \texttt{R} &&&\\
\makecell[tl]{ResearcherBench \\\citep{xu2025researcherbench}}                     & Sci:AI         & synthetic queries & \texttt{\small WS}                                                  & \texttt{\small RCC,CP, CR }           &   $\checkmark$     & - &  $\checkmark$* & \texttt{\small OA} & \texttt{\small R}  &&&\\
\makecell[tl]{FINDER\\\citep{zhang2025fargenuinelyusefuldeep}}               & Gen          & curated queries            & \texttt{\small WS}                                     & \texttt{\small RC,CP, CR}              &   $\checkmark$ & - &  $\checkmark$ & \texttt{\small MA,OA} & \texttt{\small R}  &&&\\
\midrule

\multicolumn{13}{l}{\tiny \makecell[tl]{
\textbf{COLUMNS: \hspace{2.3mm}} \textit{\textbf{Input Source:}} Sources of input (e.g., queries, topics) in the benchmark. "Curated" and "written" queries stand in for to "expert-curated" and "expert-written", respectively.\\ 
\hspace{14mm} \textit{\textbf{Expert Eval:}} Does the evaluation  use comparison to expert human annotation?;  \ \  \textit{\textbf{PPR:}} Pairwise preference ranking used for meta-evaluation; \\
\hspace{14mm} \textit{\textbf{Last 3 columns:}} For papers evaluating on pairwise-preference ranking (PPR) only, how many report-generating systems are compared (\textbf{\# Sys Comp.}), \\
\hspace{17mm} what is their reported inter-annotator agreement (\textbf{IAA}), and what is their reported human-model overall agreement (\textbf{OA})? \\
\textbf{ABBREVS: \hspace{3mm}} \textbf{\textit{Retrieval Sources:}} \{PI: Paper index, WI: Webpage index, WS: Web Search, GT: Groundtruth docs\}; \\ 
\hspace{14mm} \textbf{\textit{Metrics Used}:} \{FC: Fact Check,  RC: Rubric Coverage, RCC: Rubric Coverage restricted to content, CP: Citation Precision, CR: Citation Recall, AQ: Answer Quality\};\\ 
\hspace{14mm} \textit{\textbf{Metric-wise Meta Evaluation:}} human eval specifically designed for \{$\checkmark\checkmark$: all metrics\, $\checkmark$: a subset of metrics, $\checkmark$*: a subset of metrics, only overall score reported\};  \\
\hspace{14mm}  \textit{\textbf{Reported Comparisons:}} \{SC: system correlation, IC: instance correlation, OA: overall agreement, MA: metric-wise agreement, MC: metric-wise correlation\};\\
\hspace{14mm}  \textit{\textbf{Qs Assigned:}} \{R: randomly assigned Qs, C: expert chosen Qs, A: expert authored Qs\}\\
}}\\ \bottomrule
                                                 
\end{tabularx}
\caption{We summarize benchmarks that evaluate scientific or research-oriented systems and represent the subset that are most closely related to \sqaeval. Ordered alphabetically by author.} %\aps{"Reading the description about PPR separate from the rest of the Meta eval description later is a bit jarring"} \jena{aman, could you say more about PPR thing?}}
\label{tab:related-works1}
\end{table*}

\section{Current Status of Long-form Scientific Report Evaluation}

The past year has seen a surge of research on evaluating long-form, retrieval-augmented deep research  systems with benchmarks designed to assess both the quality of generated reports and how well they are grounded in external sources through citations.
%The past year has witnessed an explosion of research targeting the evaluation of long-form, retrieval-augmented deep research systems. These benchmarking efforts seek to assess the quality of generated reports with the assumption that high-quality deep-research reports must be grounded in external sources, and that evaluation should explicitly assess both the content of the answer and its use of citations. 
Given a query, these efforts grade system reports along dimensions of answer quality and citation support. 
We identify a total of 17 benchmarks that evaluate scientific or research-oriented systems, and their corresponding meta-evaluations. These represent the subset that are most closely related to \sqaeval and our present meta-evaluation. Table~\ref{tab:related-works1} summarizes 12 benchmarks that use agreements with human scores for their meta-evaluations (See Table~\ref{tab:related-works-complete} in Appendix for an expanded version).

\paragraph{Evaluation Benchmarks.} 
The listed benchmarks use an LLM-as-a-judge evaluation framework, in which an evaluator LLM processes the original query and the system-generated report, and computes metric scores that capture different aspects of response quality. 

Content quality is generally assessed via factual check using a set of \textit{ground truth facts} (abbrev. FC in Table~\ref{tab:related-works1}) \cite{coelho2025deepresearchgym,Fan2025UnderstandingDV,patel2025deepscholarbench,pradeep2024trecrag, Wang2026DeepResearchEvalAA}; or by evaluating coverage of key aspects expected in high-quality answers via \textit{evaluation rubrics} (RC, RCC). 
For some works, these key aspects encompass a range of content quality measures like organization and attribution (RC)  \cite{Bosse2025deepresearchbench, du2025deepresearch, Li2026DeepResearchBI, wang-etal-2023-automated, Wang2026DeepResearchEvalAA, zhang2025fargenuinelyusefuldeep}, while others restrict to content coverage only (RCC) \cite{bragg2025astabenchrigorousbenchmarkingai, asai2024openscholar, yifei2025researchqa, xu2025researcherbench}, with separate evaluations for style and citations.

These works\footnote{With the exception of Trec Rag 2024~\cite{pradeep2024trecrag}, which specifically focuses on nugget (fact) detection.} include claim verification measured in terms of \textit{citation recall} (CR)---the fraction of claims in the report that are backed by evidence from the cited sources; and \textit{citation precision} (CP)---the fraction of cited sources that support their associated claims.
Unique to \sqaeval is the \textit{answer relevance} (AR)---quantified as the proportion of relevant passages in a report. Further details are in §\ref{sec:sqaeval}.

Beyond these works, benchmarks also assess the quality of research-based generated outputs with differing evaluative objectives such as clinical conversation \& reasoning \citep{Arora2025HealthBenchEL, Chen2025MedBrowseCompBM, Gupta2024OverviewOT}, hypothesis generation \& ideation \citep{moussa2025scholarevalresearchideaevaluation}, and deep-research reasoning \& planning \citep{gupta2025deepsearchqa}.

\paragraph{Meta-Evaluation.} 
Most current meta-evaluations rely on expert annotations, most commonly through \textit{pairwise preference ranking} (abbrev. as PPR in Table~\ref{tab:related-works1}), where expert annotators compare two system outputs for the same query and indicate their overall preference. Performance is typically reported as system-level agreement (SC) or correlations between model scores and aggregate "gold" human preferences (OA). Limited number of works include instance level correlations (IC) or fine-grained evaluations that target the design of the metrics in the benchmark, often reported as agreement(MA) or correlation (MC). %\dd{minor, but if we don't *need* the distinction between MA and MC, it would be a more parsimonious story to collapse them into one thing.}  

In addition to these aspects, our meta-evaluation extends beyond standard practices by including comprehensive metric-specific human annotations and exploring expertise-targeted evaluation via controlled question assignment ("Q Assigned") to investigate the effects of annotator expertise and query authorship. 
As we will show in this paper metric-specific comparisons enable interpretable insight into how well individual evaluation aspects measure their intended targets. This approach also helps disentangle successes from failures of the evaluation along dimensions of judgment not captured by any single metric.
%This enables systematic analysis of the successes and failures of current meta-evaluation practices, and how they can be refined for scientific deep-research contexts. \dd{It's hard to tell how the particular additional measurements mentioned in the first sentence lead to the  (broad) benefits in the second.  Can we make the stated benefits more specific?}

Finally, for papers that report on pairwise-preference annotation results in Table~\ref{tab:related-works1}, we include the reported inter-annotator agreement (IAA), overall agreement score (OA), and the number of report generator system the human evaluation compares. We reference these values in \S\ref{sec:additional-findings}.

\section{\sqaeval Benchmark}
\label{sec:sqaeval}

%\dd{We need to make it more clear why we are explaining all of these things---how does this section help make our contributions?}
In this section, we provide a brief overview of \sqaeval and describe the metrics used in the benchmark in some detail as we adopt them in defining our meta-evaluation settings.
\sqaeval was recently released as a part of a suite of agentic evaluation framework called \astabench \cite{bragg2025astabenchrigorousbenchmarkingai}. 
It is designed to evaluate deep research systems, in particular, systems that use retrieval-augmented generation to produce reports that are grounded in cited literature. 
The evaluation takes as input a user query and a generated report, and assesses the report along four constituent measures of answer quality based on report content and its attributions. The final output score is an average of the four metrics, individually described below. 

\sqaeval uses LLM as judge, adopting Gemini-2.5-Flash for its strong accuracy-cost tradeoff (further explored in \S\ref{sec:pairwise_results}).
%for its favorable balance of accuracy and cost-effectiveness 
%in the metric evaluation outcomes in our experimentation, which we further explore in \S\ref{sec:pairwise_results}.
The benchmark consists of 100 real CS-domain queries in the test set and 100 queries in the dev set.
%development set from ScholarQA-CS \cite{asai2024openscholar} with updated rubrics.
The evaluation benchmarks 10 systems: 8 deep-research agent systems and 2 frontier LLMs.

\vspace{.3em} \noindent \textbf{Metrics: Answer Recall.}
This metric measures the fraction of necessary points or "rubric ingredients" covered by the report. 
Each question has an \textbf{answer rubric} that identifies the key ingredients that a correct answer must include.  These are \textit{content} rubrics that focus on the main substance and information a report should contain. This approach (i.e., RCC shown in Table~\ref{tab:related-works1}) diverges from general rubric coverage measure (RC) that includes ingredients that measure writing mechanics such as structure, clarity, and formatting.

The rubric generation is fully automated. Similar to recent TREC information retrieval competitions, \citep{10.1145/3726302.3730090}, candidate ingredients are pooled using \texttt{claude-opus-4-20250514} and are merged subsequently. 
%\footnote{Our approach follows \citet{pradeep2024trecrag}'s extract-and-merge paradigm used for AutoNuggetizer, where LLMs are used to extract factual nuggets from retrieved passages and merged an answer key.}
% For each question, the LLM extracts key ingredients with specific details from each system's answer and labels them as "answer critical" (must-have) or "valuable" (nice-to-have). The LLM then clusters semantically similar ingredients while preserving importance labels, creating question-specific rubrics used to measure coverage at evaluation time. For each ingredient cluster, the LLM judge assigns scores of 0 (doesn't meet criterion), 1 (somewhat meets), or 2 (perfectly meets). The final answer coverage score is a weighted average of ingredient scores, with "answer critical" ingredients weighted twice as much as "valuable" ones. \dd{we may want to dig up and cite a recent TREC competition, I think in 2024 or early 2025, where they did something closer to exactly our design of extract-and-merge.  I'm not sure what the exact ref is.  We didn't know about it until after doing our work.} \jena{I added the pradeep paper. I also added a footnote, but if this is too much  feel free to comment it out.} \dd{I swapped out the pradeep paper for one that describes AutoNuggetizer in detail, and removed the Craswell one.  With that I think we don't need the footnote so I commented out.}
%
We validate our rubrics by comparing them against human-created and LLM-generated versions through expert annotation (details in §\ref{sec:appendix-rubric-validation}).

\vspace{.3em} \noindent \textbf{Metric: Answer Relevance.} This metric is defined as the fraction of paragraphs in the report that directly address the question. Given the full report, the LLM judge is instructed to flag all irrelevant paragraphs. 
%\aps{Can probbaly be merged with answer recall as the citations metrics below?} 
%\jena{oh i didn't realize I didn't finish this section. I think it may be a bit confusing to join with the rubric bit in there as well. }
%
Our initial qualitative analysis revealed that deep research systems more commonly err by straying off-topic rather than through outright hallucination. Consistent with this finding, the \textbf{answer relevance} metric measures this error and %Complementing our content-based rubric evaluation that assesses whether necessary answer components are present and our citation evaluation that verifies that claims are correctly attested with citations (described below), answer relevance
penalizes passages that don't directly address the query according to the LLM judge. To the best of our knowledge, answer relevance evaluation is the first of its kind in evaluating a deep research system response based on relevance.  
%\dd{will we include any prompts in this paper's appendix, or just refer to AstaBench paper?  I feel like a pointer to those prompts somewhere seems important.} \jena{+1 got it; we'll have to include them for tacl in the appendix}

\vspace{.3em} \noindent \textbf{Metrics: Citation Precision and Citation Recall.} 
\textbf{Citation precision} measures the proportion of citations that fully or partially support the claims to which they are attached, and \textbf{citation recall} measures the proportion of claims that are fully supported by their accompanying citations. For each section, the LLM judge is instructed to identify a list of claims and
 %These two metrics are calculated from a unified prompt, where we give our LLM-judge a report, section-by-section, and we ask it to identify a list of claims, 
 their corresponding citations, and judge whether the claims are fully supported by the citations and which of the provided citations support the associated claim. 

%\dd{need to explain the excerpts case a bit more completely, saying most (how many on our study?) systems today provide excerpts and we use those to evaluate support.} \jena{addded. additionally, \# systems are eval'd is included in the last intro paragraph in this section. }
To evaluate claim support, the citation metrics require evidence excerpts from the cited sources, which the 8 benchmarked deep-research agent systems are capable of producing. In case only the source title is available,\footnote{The two benchmarked LLMs often return only titles. It can also happen if the particular sources' texts are unavailable to the deep-research system e.g. for copyright reasons.} a half-credit is awarded based on the LLM judgment.
% For each claim, if the LLM judge assesses that the claim is fully supported by its citations, that claim receives a citation recall score of 1.0 (or 0.5 if no snippet is provided from the cited text, 
% this can happen because the system lacks the quote feature or because the particular sources' texts are unavailable to the system e.g. for copyright reasons). Otherwise, the claim receives a score of 0.
The final citation recall measure is an average over claims and the citation precision is the average of these scores macro-averaged by claim.  

%To compute citation precision, we use the LLM judge assessments of whether a citation provides at least partial support for its associated claim.  If yes, the citation receives a score of 1 (or 0.5 if it lacks a quote), otherwise it gets a score of 0. 
%\aps{This can probably be cut and the following text added to the end of the previous paragraph} 
%\jena{sounds good!}
%\jena{Brief discussion about complexity of 0.5credit assignment} \assign{Dany?}

\section{Meta-Evaluation via Comparisons to Human Pairwise Judgments}
\label{sec:meta-eval}

\begin{figure}[t]
  \centering
  \includegraphics[width=\linewidth]{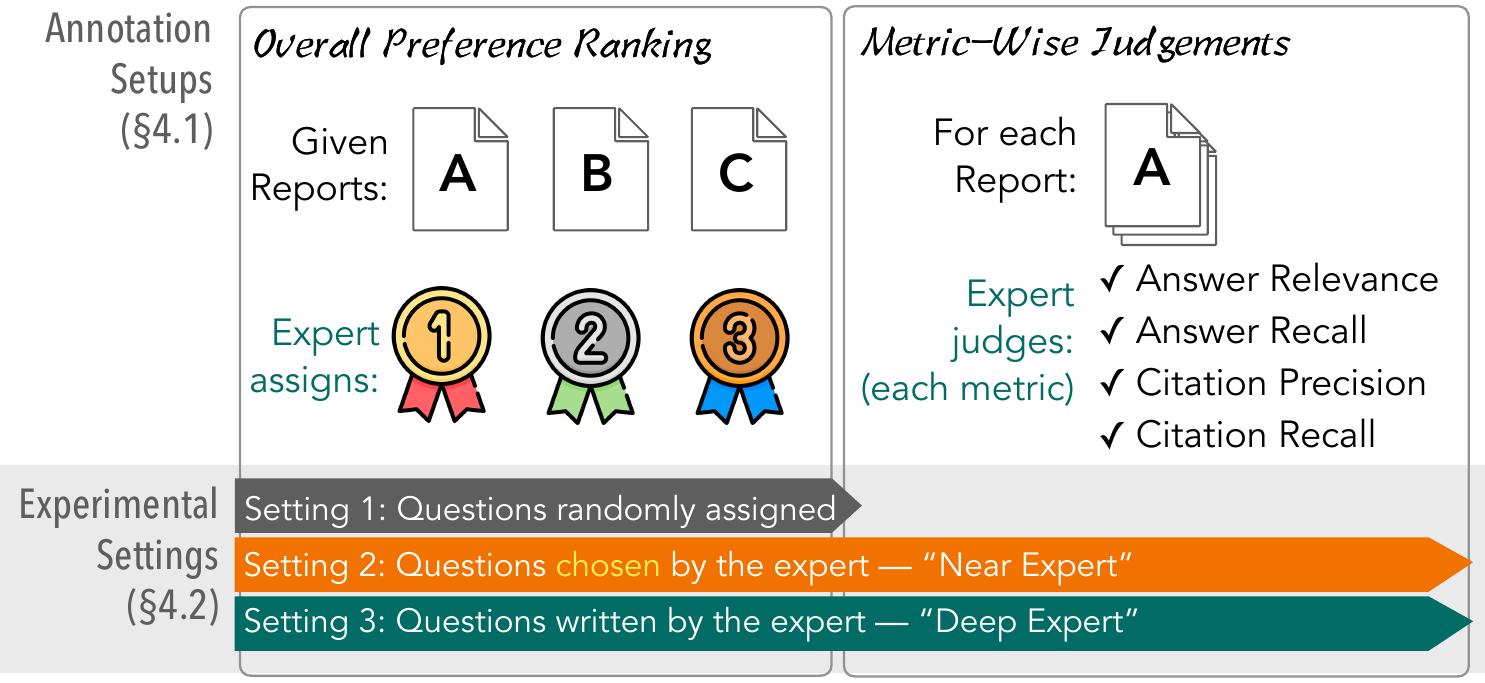}
  \caption{Our experimental settings for meta-evaluation of \sqaeval. We investigate meta-evaluation that assesses agreement between evaluation score and expert preference ranking (Setting 1). We also investigate settings (Settings 2 \& 3) that compare evaluation scores against metric-wise judgment annotation and control for annotator expertise level.}
  \label{fig:exp-setup}
  \vspace{-3mm}
\end{figure}

% limitations
%https://arxiv.org/abs/2410.20266

Pairwise human judgment is generally favored in long-form generation meta-evaluation
for providing a cost-effective, yet competent approach that yields a consistent ranking of systems or outputs \cite{liu2025aligninghumanjudgementrole}.
%because it sidesteps the limitations of often conflicting nuanced human judgments and provides a cost-effective, yet competent approach that yields a consistent ranking of systems or outputs \cite{liu2025aligninghumanjudgementrole}. 
Relative preference rating sidesteps the limitations of often conflicting nuanced human judgments by relieving the annotators from the need to understand and apply precise definitions of individual criteria, taking advantage of fundamentally comparative nature of human decision \cite{Stewart2006DecisionBS}.
This is particularly useful for setups like deep research, where the absence of a single “correct” answer makes absolute scoring challenging \cite{chiang2024chatbotarenaopenplatform, Zhao2025SciArenaAO}. 
Relative judgments thus help circumvent these issues, rendering evaluations easier and more reliable for annotators. 
Relative comparisons can also surface fine distinctions that may be missed in a predefined, absolute point-wise judgment setting \citep{Yan2022HumanPA, Clarke2020AssessingTP}.

%pairwise comparison or ranking may better align with the cognitive processes underlying human evaluation.

% @article{Stewart2006DecisionBS,
%   title={Decision by sampling},
%   author={Neil Stewart and Nick Chater and Gordon D. A. Brown},
%   journal={Cognitive Psychology},
%   year={2006},
%   volume={53},
%   pages={1-26},
%   url={https://api.semanticscholar.org/CorpusID:9122872}
% }

% Normative decision theories assume that people have
% stable and consistent preferences regardless of how the
% preferences are elicited.

%\dd{"overcomes the limitations" seems too strong---do we just mean it is more reliable than Likert scores?} \jena{I think what I'm trying to say here is that: Since pairwise judgment captures wholistic preference, you are sidestepping the problem of specific, more attribute/criteria-based, judgments that are susceptible to the subjectivity because of its specificity, nuance, and interpretative differences (between say those who wrote the questions/criteria and the annotators). Would sidestep work?} Re preference-based, I think it may just be helpful to explain more: 
%giving a relative preference rating does not require annotators to understand and apply definitions of individual criteria and likert scale  values.  And then (if there’s empirical support for this and I think we had some citations to that effect?) say that this can result in easier or more reliable annotations.

Thus, following the standard long-form generation meta-evaluation standards in our field, we design human pairwise assessments and compare them to \sqaeval score (also referred to as ``model score"; see Figure~\ref{fig:exp-setup}):
%We conduct meta-evaluation in two primary settings:
% We do two things:
(i) We collect \preferencerank{overall preference judgments} where experts rank answer reports according to their preferences and analyze the results at system and instance level% 
%\dd{Either here or somewhere I think we need to explicitly define what system-level and instance-level mean.}
.\footnote{Instance-level agreement compares each evaluation score to annotator label, before aggregation.  System-level agreement compares aggregated evaluation scores, averaged across a system’s instances, to aggregated human annotations.}
%Instance-level agreement assesses at specific instance before any aggregation. System-level is aggregate agreement--average of averages across instances in the set.} 
(ii) We generalize beyond overall preference-based meta-evaluation and conduct systematic human annotation over 
each of the four \metricwise dimensions evaluated in \sqaeval.
% of all dimensions of evaluations represented by each of our metrics. 
% \sqaeval includes four metrics. 
% We conduct pairwise annotation that is specifically designed to address each of the dimensions of quality the four metrics represent. 
(iii) We further control for the expertise level of our annotators to assess its effect in human-LLM agreements for the above settings.
Together, these settings allow us to examine the role of each component of our LLM-judge evaluation and assess its contribution to overall evaluation accuracy and its correspondence to various aspects of human judgment.

%Assessing answers to deep research questions requires domain expertise. \todoit{motivate expertise control with prior work}.

%3.

% This section presents our experimental setup.
% and results of the pairwise assessments. 
Below, we describe our annotation task setups (\S\ref{sec:annotation_setup}) and the experimental settings (\S\ref{sec:experimental_settings}) used across all human evaluation experiments conducted as part of our meta-evaluation, then summarize our principal findings in the following section (\S\ref{sec:pairwise_results}).

\subsection{Annotation Setup}
\label{sec:annotation_setup}

\paragraph{Annotators.} 
We source from a pool of 5 experts (4 Ph.D. and 1 M.S.) through the annotator recruitment platform Upwork. Four are CS experts, while one specializes in Math with attested CS annotation history. They were paid \$31/hour on average.

\vspace{.25em}\noindent\textbf{Overall preference ranking.}
For \preferencerank{overall preference ranking annotation}, our experts are presented with a question and reports generated by three different systems, sampled from a pool of six systems: \sqaeval \cite{bragg2025astabenchrigorousbenchmarkingai},
%singh2025ai2scholarqaorganized}, 
\sqaeval-SFT,\footnote{\sqaeval-SFT is a Qwen3-8B model finetuned on QA pairs collected from production \texttt{SciQA}.} Elicit  \cite{elicit2024}, OpenAI-DR \cite{openai2025deepresearch}, Perplexity \cite{perplexity2024}, and Storm \cite{shao2024assistingwritingwikipedialikearticles}. For each 3-way comparison, the annotators are first asked to read each report individually. After reading the reports, they rank the responses by assigning gold, silver, and bronze medals based on overall answer quality, with ties permitted.
%\dd{nothing we need to act on right now but, is there any signal to the absolute scores, e.g. did they ever assign all bronze vs all gold?  I believe we're only using the relative scores. \jena{It happened for a few instances in the initital annotation rounds. We subsequently barred them from assigning 3 way ties as it felt like those instances that were happening could have been distinguished, based on manual examination of annotator comments, other scoring etc.}}. 
In assigning the rankings, they are specifically asked to use their own preference judgement. 
%They are also asked to prioritize the following four aspects of the report, which mirror our four evaluation metrics: whether the generated answer directly addresses the question, maintains relevance throughout, provides adequate citation coverage, and uses citations that meaningfully support its claims.  In later annotations, we ask the annotators to explicitly label reports according to those metrics, but in this overall preference ranking they are only provided as general guidelines .
%\aps{Can we be explicit here and mention that to get a sense of how closely the overall preferences align with the individual metrics, we mention in the instructions to prioritize those aspects, and then link to the instructions?} \jena{what do you mean by "sense of how closely pref align with individual metrics? You mean like Fig 4 and 5?}

The average completion time for each 3-way comparison was 45-50 minutes. Annotation guidelines, illustrative procedure, and screenshots of the interface used is included in §\ref{sec:appendix-guidelines}.

\vspace{.25em}\noindent\textbf{Metric-wise judgments.}
% We implement an annotation set up that modifies the overall pairwise preference to 
For \metricwise annotation, we explicitly introduce metric-specific annotations for each report.
% \textit{before} annotators assign overall preference ranks. 
% This setup is designed to closely approximate the four metrics in \sqaeval. 
As with overall preference annotation, each task consists of a question and reports from three systems 
%\dd{We need to be careful about terms.  We use "model" to refer to the output of an eval ("model score"), but here we are using it to refer to what elsewhere we call systems.  Probably we should refer to deep research systems as "systems" always.  And for evals, I think I'm okay  with using the term "model" and "model score" but we should say that "we refer to these scores as "model scores"" or something like that}. \jena{added to the 2nd intro par in this section}
For each report, the annotator completes three sub-annotation tasks corresponding to answer relevance, answer recall, and citation recall \& precision. 
% Each of these sub-tasks is completed after each report and before the overall preference based ranking. 

For judging \emph{answer relevance}, annotators use checkboxes to flag irrelevant passages during reading.\footnote{The annotation instructions use the contents of the prompt used for our evaluator model.} 
\emph{Answer recall} is evaluated as a single question: ``Did the report cover all aspects of the answer you expected for it to cover?'' with answer choice on a 5-point Likert scale. For citation evaluation, the experts pick 5-8 claims central to the report 
%
%
%=====> return to this
%\dd{Nothing to change at this point, but I only just now realized that we asked them to pick central claims, whereas the metrics average over all claims.  I wonder if this matters, e.g. it could be that central claims are better-supported than random ones and so the citation scores veer lower (especially for longer reports, with more non-central claims) than the eval for this reason.} 
%
%
and  assess if the claims were fully/partially supported by the associated citation(s) (i.e., \emph{citation recall}) and surface the number of citations that were relevant to the claim given its associated citations (i.e., \emph{citation precision}). For this task, the experts are asked to carefully consult the evidence provided by the report.\footnote{For each report we collect four additional assessments on a 5-point scale:\textit{informativeness}, \textit{background}, \textit{organization}, and \textit{flow} (annotation step is marked with asterisk in \S\ref{sec:appendix-guidelines}). \label{footnote:qualitative-eval}}
Following completion of these three sub-annotation tasks for each of the three answers, the annotator provides overall preference ranking as in the prior task setup. 

The average completion time for each 3-way comparison in this metric-wise plus overall setting was 90 minutes. The screenshots of the interface used is included in §\ref{sec:appendix-guidelines}.

\vspace{.25em}\noindent\textbf{Result tally.}
Our annotations result in 3-way rankings over the answering reports. We convert these results (overall and metric-wise) into pairwise judgments between each of the three pairs of report (Figure~\ref{fig:score-agreement-calc} in §\ref{app:expt_details}). The conversion of annotated values to scores is detailed in §\ref{sec:appendix-score-conversion}. 
% For metric-wise judgments, the annotated values are decomposed, again, into three model pairwise pairs. 
Since human judgments include ties, we use thresholds to determine ties in \sqaeval's numeric scores during conversion to ensure a level of fairness when the scores are close enough to be considered ties. The process of tie thresholding is detailed in §\ref{sec:appendix-thresholding}. \metricwise overall score is the average of the expert's individual metric scores, calculated at instance-level, to serve as comparison to model overall score.
%We further investigate tie thresholds for LLM-judge evaluators to unders and their impact on meta-evaluation results
% ; the expected random chance (by default) is over three labels. We compare model scores against the collected human pairwise judgments for both pairwise preference rankings and 

\begin{table}[!t]
    \centering
    \scriptsize
    \begin{tabular}{@{}lcc@{}}
        \toprule
        ~ & All Data & Expert-\\ 
        &(IAA: 55.0\%)& Agreed\\ \midrule
        \textbf{System-level} Corr ($\tau$-b) & 0.40 & 0.60 \\
        $\hookrightarrow$ w/o Elicit & 0.70 & 1.00 \\ [0.5mm]
        \textbf{Instance-level} Corr ($\tau$-b) & 0.25 & 0.36 \\
        \bottomrule
    \end{tabular}
    \caption{System and sample-level correlations (Kendall $\tau$-b) and Human IAA. Our system-level $\tau$s range from moderate-to-strong. The strong correlations are not reflected at instance-level comparison.}%\dd{It may be that we don't have space, but these tables are not self-contained.  In a perfect world, would explain more---what are the two columns, what is without elicit.  Also ideally, every fig or table has a single-sentence take-home message at the end, the most important thing you want the reader to take away from the fig/table.}\jena{I agree with you. We can do this for the arxiv where we don't have page limit.}}
    \label{tab:score-correlation}
\end{table}

\subsection{Experimental Settings}
\label{sec:experimental_settings}

\paragraph{Setting 1: \textcolor{darkgray}{\preferencerank{Overall Preference}; Random assignment.}}

We conduct the overall pairwise preference task over questions in our test and dev set. The questions were \textit{randomly} selected for each annotator. We collect 3-way comparisons in the overall pairwise setting for 100 dev instances and 100 test instances, resulting in a total of 600 pairwise judgments. All comparisons are doubly annotated to compute inter-annotator agreement (IAA). We compare LLM-judge results against collected human preferences, reporting Kendall's $\tau$-b (Table~\ref{tab:score-correlation}) and pairwise agreement percentage (Table~\ref{tab:score-agreements}).\footnote{Cases in which one annotator declares a tie and the other does not are counted as a \textit{disagreement} for our IAA. If ties are relaxed by awarding half-credit in such cases, our IAA improves from the reported 55.0\% to 82.9\%. The choice to use thresholds to calculate model's winner was based on this decision to retain human ties. \label{footnote:iaa} }

%We validate \sqaeval by matching system's results against human expert's  preferences given two pairs of reports, and quantify the ranked relationship between \sqaeval's average model scores and expert evaluations using Kendall’s tau-b correlation. 
%
%We compare LLM-judge scores against the collected human pairwise preference ranking. 
%Results are summarized in Table~\ref{tab:pairwise}.  

\vspace{0.3em} \noindent \textbf{Setting 2: \textcolor{nearexp}{\metricwisecap; Near-expert assignment.}} 
In this setting, experts are asked to select questions aligned with their own expertise by reviewing all questions from the test set \emph{not} assigned during previous rounds of annotation and categorize each as being \textit{within}, \textit{near}, or \textit{outside} their area of expertise. For each annotator, we discard the \textit{outside} category, and select 5 questions per expert giving priority to the \textit{within} label. Each expert then provided judgments on reports for their five questions in the metric-wise annotation task setup.
For this setting, we only employ the four annotators with expertise in CS as all our questions are from the CS domain. In this manner, we collect 60 pairwise annotations across 20 questions. The results are summarized in Table~\ref{tab:controlled-setup}.

\vspace{0.3em} \noindent \textbf{Setting 3: \textcolor{deepexp}{\metricwisecap; Deep-expert assignment.}}
To ensure even stronger alignment with the annotator’s specific expertise, we conduct the same experiment as Setting 2 but asking each expert to {\em write} five \emph{new} questions on a topic that they are intimately familiar with, ideally relating to questions from their current or past research. We additional guidance and instructions to ensure question quality. 
%For the annotation task, each expert then judges answers for the questions they wrote. 
By having experts evaluate answers to their own questions, we ensure that evaluations are grounded in deep domain knowledge and a complete understanding of what is required to answer the question, thereby reducing the potential noise of surface-level evaluations.
%\dd{it is not made clear necessarily that we expect the annotators to be more expert on these questions compared to the last setting.  Is this what we believe, and if so why?} 
%
The reports are formatted and the rubric generation pipeline (described in \S~\ref{sec:sqaeval}) is used to compile question-specific rubrics. We employ all 5 annotators in this setting, collecting 75 pairwise annotations across 25 questions.\footnote{For deep-expert setting, we substitute Elicit with FutureHouse's Falcon (https://edisonscientific.gitbook.io/). We were unable to obtain access to Elicit for this experiment.} We summarize our results in Table~\ref{tab:controlled-setup}.

\section{Principal Findings}
\label{sec:pairwise_results}

\begin{table}[!t]
    \centering
    \scriptsize
    \begin{tabular}{@{}llcc@{}}
        \toprule
        Human Score & Model Score & All Data & Expert-Agreed\\ \midrule
        \multirow{4}{*}{\preferencerank{\scriptsize \textbf{\makecell[l]{Overall\\preference\\ranking}}}} & \cellcolor{lightgray} Overall Score &  \cellcolor{lightgray} 51.6\% & \cellcolor{lightgray}  62.4\%  \\  
        & Answer Relev. & 34.5\% & 35.2\%  \\ 
        & Answer Recall& 47.3\% & 54.3\%  \\ 
        &Citation Prec. & 42.3\%  & 59.0\% \\
        &Citation Recall & 41.9\% & 58.4\%  \\ 

        \bottomrule
    \end{tabular}
    \caption{Pairwise agreements between expert overall preference ranking vs. model scores. Our overall score agreement approaches IAA levels, however, individual metric comparison to human preference show lower levels of agreement.}
    \label{tab:score-agreements}
\end{table}

% DO NOT ERASE BELOW
% \begin{table}[!t]
%     \centering
%     \scriptsize
%     \begin{tabular}{@{}llrrrr@{}}
%         \toprule
%         & ~ & \multicolumn{2}{c}{\makecell{All Data}} & \multicolumn{2}{c}{\makecell{Expert-Agreed}} \\ 
%        Human Score & Model Score & corr. & agree. & corr.  & agree.\\ \midrule
%         \multirow{4}{*}{\makecell[l]{preference\\ranking}} & \cellcolor{lightgray} Overall Score & \cellcolor{lightgray} 0.25 & \cellcolor{lightgray} 51.6\% & \cellcolor{lightgray} 0.36  & \cellcolor{lightgray}  62.4\%  \\  
%         & Answer Relev. & -0.07 & 34.5\% & -0.08 & 35.2\%  \\ 
%         & Answer Recall& 0.16  & 47.3\% & 0.23 & 54.3\%  \\ 
%         &Citation Prec. & 0.22 & 42.3\%  & 0.27  & 59.0\% \\
%         &Citation Recall & 0.23 & 41.9\%  & 0.27 & 58.4\%  \\ 

%         \bottomrule
%     \end{tabular}
%     \caption{Pairwise agreements: expert pairwise preference ranking vs. model scores.}
%     \label{tab:instance}
% \end{table}

We organize our results into four principal findings. 

\vspace{1em}
\begin{mdframed}[backgroundcolor=gray!20] 
Finding 1: \textbf{\preferencerank{Overall preference ranking} is a useful meta-evaluation strategy for assessments of overall system performance, but is not suited for assessing performance at instance- or metric-level.}
\end{mdframed}

\vspace{1em}

\noindent As shown in \textbf{Table~\ref{tab:score-correlation}}, we observe a moderate system-level correlation of 0.40. Over the subset of data where all experts agree (i.e., 100\% IAA), we observe a stronger correlation of 0.60. Further, when Elicit outputs are excluded from the evaluation, the system-level correlation rises markedly to 0.70 across all pairs and reaches a wildly high coefficient of 1.00 for expert-agreed instances.
Despite high system-level correlations, the instance-level correlation is substantially weaker at 0.25 for all data, which rises to 0.36 for expert-agreed instances. As it was for system-level, instance-level relationship is affected by a systematic disagreement over Elicit instances.\footnote{Manual inspection of error cases showed a consistent pattern of disagreement between human judgments and our evaluation method when the comparison set included an Elicit-generated report, human raters tending to rank the Elicit report lower than the model score. Because we convert all reports presented to annotators in the same format in order to minimize potential stylistic bias towards any particular agent's report, the systematic human preference we observe for Elicit instances may be due in part to our edits that diverge from their intended presentation.} 

In \textbf{Table~\ref{tab:score-agreements}}, 
%shows score agreements between system scores and experts' overall preference rankings at instance-level. W
we compare expert's preferences against \sqaeval's overall score (a computed average of four metrics) and against each of the individual metrics. We observe an overall score agreement of 51.6\%, a few points shy of the IAA (55.0\%), which suggest that models agree at similar levels as human judges themselves. 
Comparing \preferencerank{preference ranking} against individual metrics, we see lower levels of score agreements: 
%citation precision and recall show strongest value at 42.3\% and 41.9\%, respectively for the full dataset and 59.0\% and 58.4\% for the expert-agreed cases.
answer recall and citation metrics show similar levels of agreement, ranging from 42-47\% for the full dataset and 54-59\% for expert-agreed cases.
Answer relevance agreement is comparatively quite low, capped at 35.2\% even for expert-agreed instances. That overall score agreement is higher than that of any individual metric suggests that the metrics together capture different facets of human judgment, complementing one another in ways that counterbalance their individual weaknesses and collectively achieving more than any single metric.

Taken together, our findings indicate that preference ranking, although widely used for its simplicity, may not necessarily reflect performance across fine-grained evaluation dimensions such as relevance or recall. Further, strong agreement at the system-level does not necessarily imply strong agreement when assessing individual reports (instance-level) or in gauging performance on a particular metric.  %Although high system agreement with human preferences is useful for verifying that the report generating system produces responses that generally align with expert expectation, \textit{agreements with human preferences should not be used to infer success in evaluating individual reports or in gauging the performance of individual metrics.}  
%\dh{is this essentially saying that we can look at aggregates and trust the number we get, but not when looking at individual instances?}\jena{trust the numbers generally that agent is overall producing reports that track with human expectations of quality. but we can't say it's doing that for any one report indivudally. }
%\vk{We should also discuss how the sample level agreement we get compares with similiar papers like researchQA and deepresearch bench and why they differ. The might be different because of ties, because of diversity in systems evaluated, etc.}

\begin{table}[!t]
    \centering
    \scriptsize
    \renewcommand{\arraystretch}{1.01}
    \setlength{\tabcolsep}{4pt}
\begin{tabular}{@{}ll|l|rr|rr@{}}
\multicolumn{3}{c}{\textbf{Human-Model Score comparison}}     & \multicolumn{2}{c}{\textcolor{nearexp}{Near-Expert}}                     & \multicolumn{2}{c}{\textcolor{deepexp}{Deep-Expert}}                    \\ \cmidrule{4-7}
\multicolumn{2}{l|}{Human Score}   & Model Score & agree. & $\tau$-b & agree. & $\tau$-b \\ \midrule 

\multicolumn{2}{l|}{\multirow{4}{*}{\preferencerank{\scriptsize \textbf{\makecell[l]{Overall\\preference\\ranking}}}}} & \cellcolor{lightgray} Overall score    & \cellcolor{lightgray} 65.0\%              & \cellcolor{lightgray} 0.50                   & \cellcolor{lightgray} 57.3\%              & \cellcolor{lightgray} 0.25                   \\
&   & Answer relev. & 31.7\%              & 0.00                   & 22.7\%              & -0.02                  \\
&   & Answer recall    & 25.0\%              & 0.00                  & 25.3\%              & 0.09                  \\
&  &  Citation prec. & 53.3\%              & 0.38                   & 49.3\%              & 0.22                   \\ 
&  &  Citation recall & 43.3\%              & 0.31                   & 49.3\%              & 0.29                   \\ \midrule

\multirow{5}{*}{\rotatebox[origin=c]{90}{\textbf{\metricwisecap}}} & \cellcolor{lightgray} Overall  score   & \cellcolor{lightgray} Overall score    & \cellcolor{lightgray} 66.7\%              & \cellcolor{lightgray} 0.42               & \cellcolor{lightgray} 58.7\%              & \cellcolor{lightgray} 0.33                   \\  
& Answer relev.                    & Answer relev. & 43.3\%              & 0.26                   & 54.7\%              & 0.23                   \\
& Answer recall                       & Answer recall    & 58.3\%              & 0.49                   & 60.0\%              & 0.21                   \\
& Citation prec.                    & Citation prec. & 50.0\%              & 0.32                   & 50.7\%              & 0.27                  \\ 
& Citation recall                    & Citation recall & 63.3\%              & 0.59                   & 53.3\%              & 0.43                   \\ \bottomrule
\end{tabular}        
    \caption{Results from per metric human evaluation with expertise control. We compare instance-level agreement between model scores and two human signals: (TOP) human preference, and (BOTTOM) metric-wise human ratings collected in a setup parallel to \sqaeval metrics. The results include thresholded ties (see \S\ref{sec:annotation_setup}). Overall score (for both human and model) is the computed average of the metric scores at instance-level.}
    \label{tab:controlled-setup}
    \vspace{-3mm}
\end{table}

\vspace{1em}
\begin{mdframed}[backgroundcolor=gray!20] 
Finding 2: \textbf{Explicit \metricwise annotation enables meta-evaluation at the metric level.}
\end{mdframed}
\vspace{.5em}

\noindent The results in \textbf{Table~\ref{tab:controlled-setup}} reveal a clear contrast between the ability of model-computed metrics to predict human \preferencerank{preference ranking judgements} versus their alignment with \metricwise human ratings. 

Model score agreements with \preferencerank{preference ranking} (top rows) sees a similar trend as what we found in Table~\ref{tab:score-agreements}. The overall score agreements are at 65.0\% and 57.3\% for \textcolor{nearexp}{near-} and \textcolor{deepexp}{deep-expert}, respectively) with stronger $\tau$s in comparison to metric-level agreements.
%Here the comparison against the \preferencerank{preference ranking} (top rows) is comparable to the results in Table~\ref{tab:score-agreements}, the relationship between  
%While overall score agreement is 65.0\% (for \textcolor{nearexp}{near-expert}) with a moderately high $\tau$=0.50, 
The relationship to most model metrics in contrast is surprisingly weak, often approaching chance levels with $\tau$ values near zero. 
Citation quality again fares better, overall, suggesting that experts attend to citation validity even when making broad preference judgments. 

When experts judge on \metricwise rating (bottom rows), however,
the metric-level alignment improves substantially. Across both \textcolor{nearexp}{near-} and \textcolor{deepexp}{deep-expert} settings, metrics generally exhibit higher agreement relative to the preference ranking setup. The agreements for answer recall sees highest benefit of explicit task alignment from both the agreement and correlation vantage points. Answer relevance also improves consistently, with $\tau$ values rising to a low-to-moderate range. 

Our citation annotation setup (§\ref{sec:experimental_settings}) allowed experts to select their own claims. This naturally led to mismatches with the LLM judge's claim selection. To enable more methodologically consistent comparison, we conducted a follow-up citation annotation study that controlled for claim span selection (details in §\ref{sec:appendix-citation-followup-annot}). This stronger alignment between model and human evaluation setting increased agreement to 69\% for citation recall and 75\% for citation precision, with the LLM judge typically more lenient than human assessors. Elicit's reports emerged as a consistent source of disagreement; excluding them raised agreement to 75\% and 81\% for citation recall and precision, respectively.

Our results show that \textit{explicit metric-to-metric comparisons provide interpretable insights into how well individual evaluation aspects measure what they were intended to evaluate}. %\textit{When seeking to draw conclusions about individual metrics, the appropriate reference point should be human experiments designed to yield judgments that align with each metric's intended purpose.}
%Overall agreement implicitly folds in evaluation dimensions not captured by the individual metrics (see next finding), and \textit{metric-wise agreement helps disentangle evaluator quality from gaps in metric coverage}.
Whereas overall agreement may reflect dimensions not captured by any individual metrics (Finding 4), metric-wise agreement helps disentangle whether low overall agreement is due to low agreement in existing metric evaluators, or due to missing metrics.
%\vk{Here's an additional point we might want to add: These results also show that the sub-metrics are all being judged correctly by their respective LLM judges. We probably have worse overall agreement because we are need additional sub-metrics fro capturing other things humans are using for their evaluation. So, sub-metric eval helps distinguish between bad judges for an eval and not capturing all the required aspects.} \jena{Does the last sentence cover what you are hoping to say?}

\vspace{1em}
\begin{mdframed}[backgroundcolor=gray!20] 
Finding 3: \textbf{Depth of annotator expertise has a significant effect on their assessments.}
\end{mdframed}
\vspace{0.5em}
%Deeper annotator expertise improves some judgments but introduces new sources of disagreement

 \begin{figure}[t]
    \centering
    \includegraphics[width=.9\linewidth]{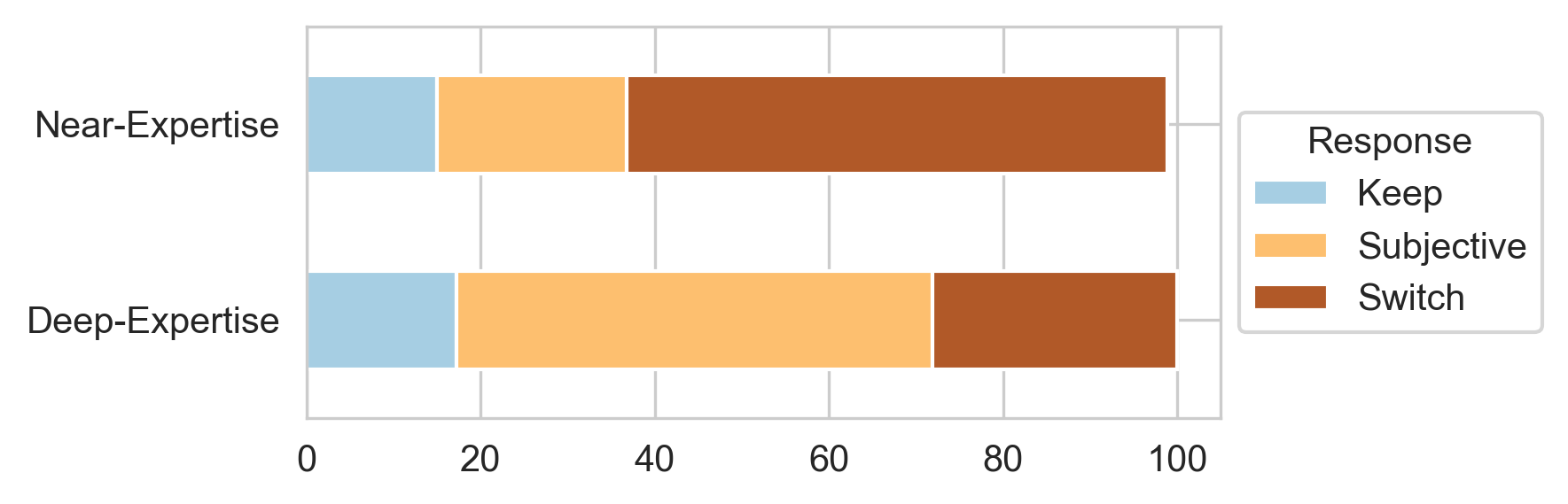}
    \caption{When evaluating answer relevance disagreements, near- and deep-experts choose to keep their own judgments at similar rates; however, when they do not, near-experts tend to defer to the LLM's judgment more often than deep-experts.}
    \vspace{-2mm}
    \label{fig:precision-disagreement-adjudicated}
    \vspace{-2mm}
\end{figure}

\noindent Comparing \textcolor{nearexp}{near-} and \textcolor{deepexp}{deep-expert} settings in \textbf{Table~\ref{tab:controlled-setup}}, we observe that the level of annotator expertise affects each \metricwise evaluation differently. Answer recall and citation precision see minimal effect. Citation recall shows a drop in agreement, suggesting that more in-depth knowledge of relevant literature leads the annotators to have more specific expectations over how extensively a claim should be supported. %, which the models do not fulfill. 

Deep-expertise sees its strongest effect in the answer relevance assessment: when we control for a tighter expertise alignment, the agreement increases from 43.3\% (\textcolor{nearexp}{near-expert}) to 54.7\% (\textcolor{deepexp}{deep-expert}), narrowing the human-model disagreement gap. 
%We posit that deep-expertise evaluation may be necessary for metrics like answer relevance where reliable evaluation requires access to detailed or nuanced understanding of the topic in question. \dd{given that the tau goes the other way for this comparison, I'm not sure we should emphasize this point.  Perhaps remove and just make the more coarse-grained, deep-expert agrees less in general than near-expert?}
% This is possibly because without the in-depth expertise assessing the relevance of nuanced details can be difficult for humans. 
To better understanding of the differences between the two settings, we conduct a follow up experiment (Figure \ref{fig:precision-disagreement-adjudicated}; details in \ref{sec:appendix-relevance-adjudication}), where the experts are asked to review answer relevance instances where LLM judge disagreed with their own answers by choosing to keep their own answer, to switch to the model answer (given model rationale), or to accept the model answer as a subjective but viable choice. There, we find that experts choose to keep their answer at similar rates in near- and deep-expert settings (15-17\% of the time), but they tend to defer to the model answer significantly more often in the near-expert than in the deep-expert setting. Instead of switching, deep experts generally view differences as subjective.

It is reasonable to expect that that annotators with greater expertise would reduce annotation noise, leading to higher agreement with the evaluation. However, the opposite appears to be true: the correlation is consistently higher in the \textcolor{nearexp}{near-expert} setting than in the \textcolor{deepexp}{deep-expert} setting.
%Since the deep-expert setting relies upon experts' extensive topic-specific knowledge, it follows that the evaluation should yield a more correct, consistent, and reliable assessment than near-expert evaluations that lacks fine-grained expertise. Counterintuitively, however, we observe that the $\tau$ for deep-expert setting is consistently lower than the near-expert even in cases where model-expert agreement rates are higher. This means, additional agreement achieved in deep-expert setting is not helping to disambiguate ranking. 
Broadly, this suggests that the LLMs possess sufficient domain knowledge to be instructed to serve near-expert judges and are capable of estimating how a general researcher might assess a report. However, it is yet incapable of mirroring deep-expert assessment capabilities and cannot capture the characteristics of a deep-expert evaluation. In other words, \textit{our results indicate that an LLM judge aligns more closely \textcolor{nearexp}{near-expert} than with the \textcolor{deepexp}{deep-expert}}.

\vspace{1em}
\begin{mdframed}[backgroundcolor=gray!20] 
Finding 4: \textbf{Observed subjectivity in human evaluation underscores task difficulty. }
\end{mdframed}
\vspace{0.5em}

\noindent Notably, our reported IAA rate is 55.0\%, meaning that experts agree on a label on about half of the evaluated instances, which speaks of the difficulty of the task and the inherent subjectivity of judging reports even at expert level setting. This raises the question: where is the subjectivity arising from? 

\begin{figure}[t]
  \centering
  \includegraphics[width=.9\linewidth]{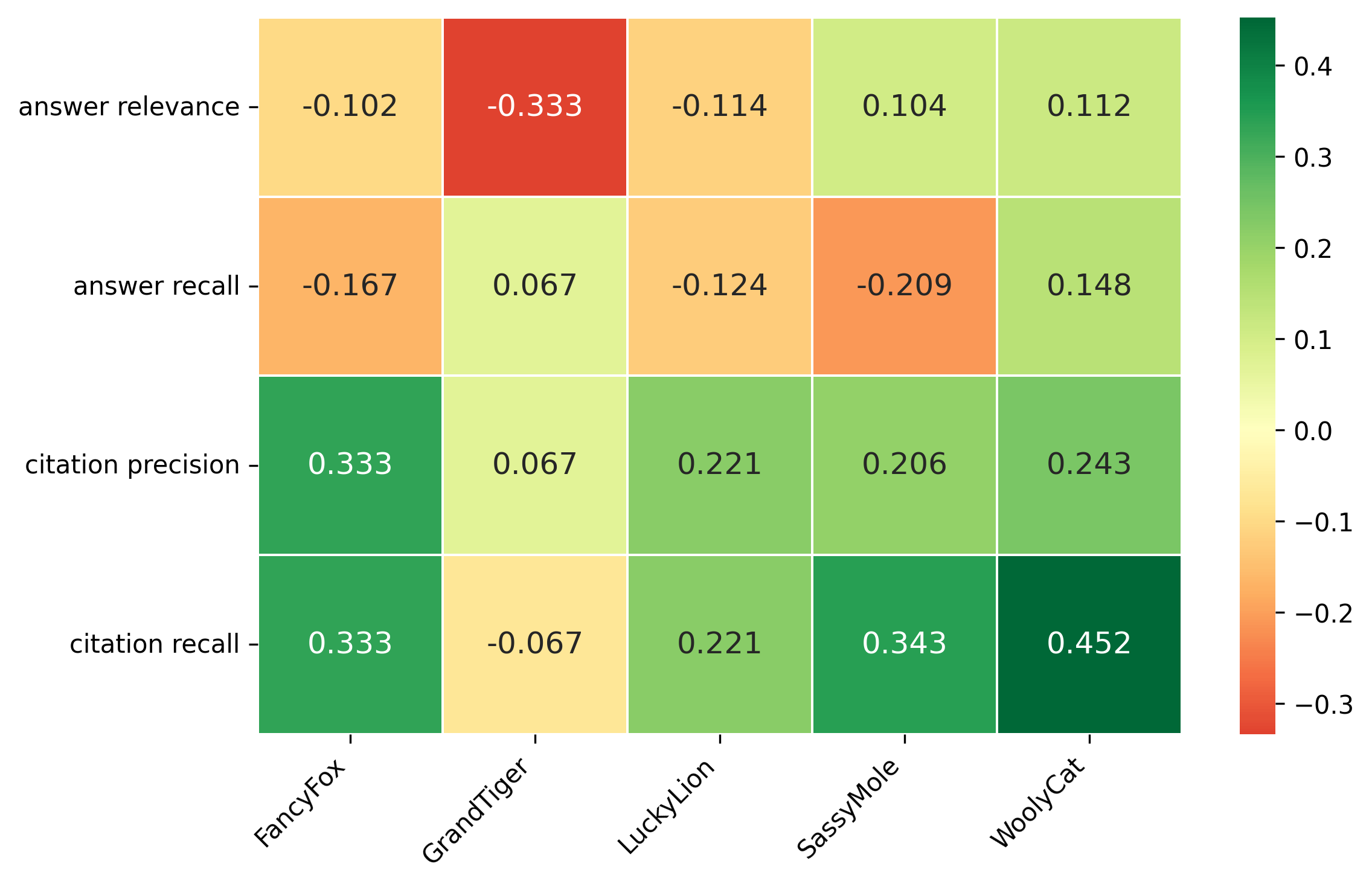}
  \caption{Expert preference ranking vs. model scores. Model score alignments with expert preferences vary substantially across annotators.}
  \label{fig:byannot-pref-vs-model}
    \vspace{-5mm}
\end{figure}

If we compare an expert's \preferencerank{preference rankings} and the model scores (Figure~\ref{fig:byannot-pref-vs-model}) from our \metricwise results (Table~\ref{tab:controlled-setup}), we observe that model scores' alignment with an expert's preferences vary substantially across annotators. Although the correlation seems generally positive with the citation metrics (with \textit{GrandTiger} as a notable exception), no single metric uniformly captures how experts assess overall quality.

\begin{figure}[t]
  \centering
   \includegraphics[width=.9\linewidth]{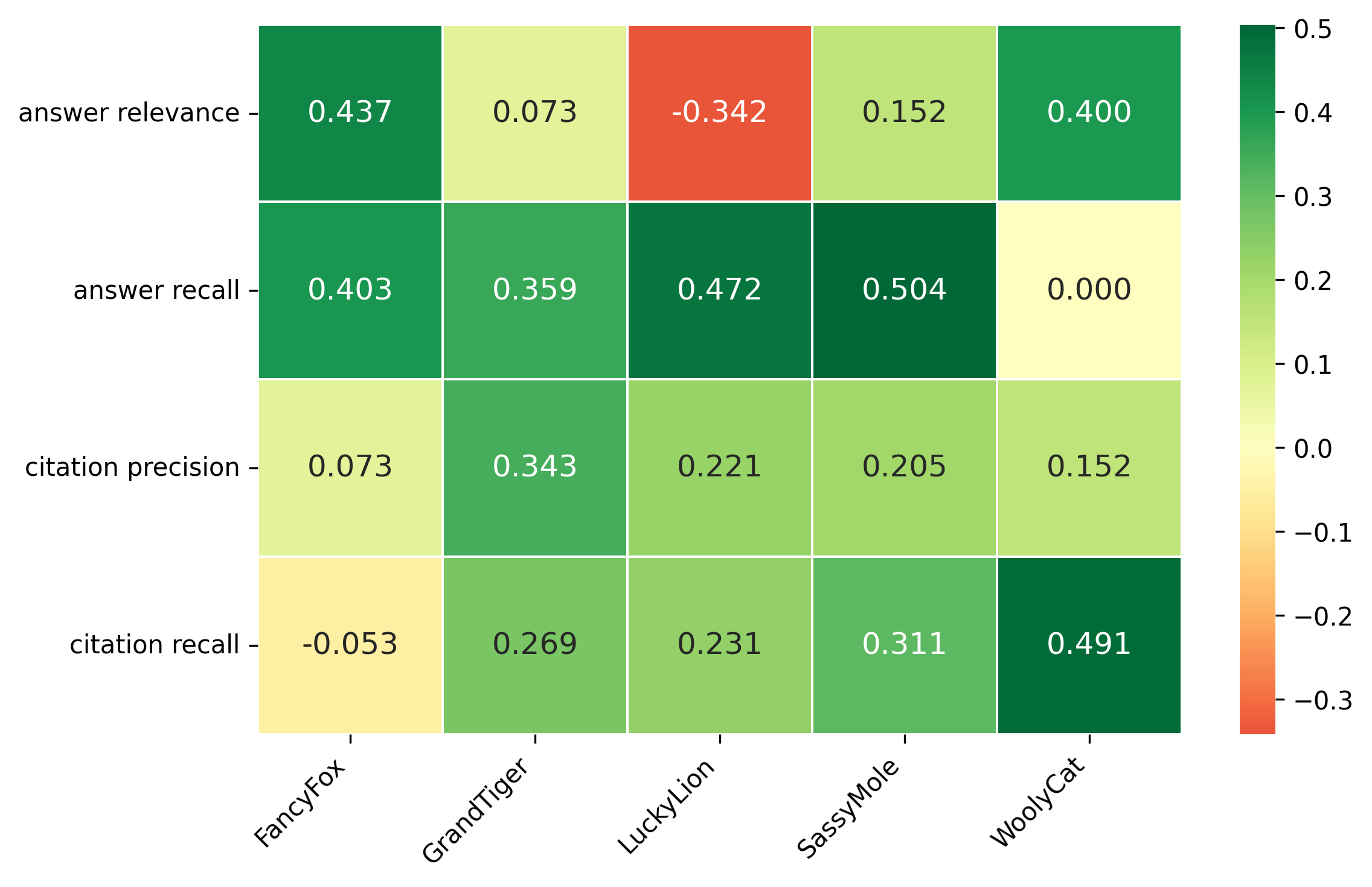}
  \caption{Expert preference ranking vs. expert's metric-wise scores obtained in annotations settings 2 \& 3. The expert preferences alignments with their \textit{own} scores vary substantially across annotators.}
  \label{fig:byannot-pref-vs-self}
    \vspace{-5mm}
\end{figure}

Our results strongly indicate that experts differ in their internal calibration of answer quality. A comparison between an expert's \preferencerank{preference rankings} and \emph{their own} \metricwise judgments obtained in Settings 2 and 3 (Figure~\ref{fig:byannot-pref-vs-self}) shows expert-specific idiosyncracies. For instance, \textit{FancyFox}’s preferences align most strongly with their own assessments of answer relevance and recall, whereas \textit{WoolyCat} places substantially higher weight on citation recall and answer relevance. In fact, these subjective expectations of answer quality extend to other aspects of writing we do not consider in \sqaeval. Analysis of the additional qualitative assessment we collected (see footnote \ref{footnote:qualitative-eval}) also suggest external factors filter into each annotator's preferences (Figure~\ref{fig:byannot-pref-vs-qualitative}) at varying levels of importance. 

The divergence in how experts weigh different quality dimensions, despite using the same evaluation criteria, suggests that interpretive differences are fundamental to the assessment task. In practice, each expert appears to calibrate what constitutes a "good" answer according to their own internal standards—standards that differ not just in emphasis but in their underlying conceptualization of quality. These findings suggest that \textit{subjective variation in human preference judgments, in a large part, stems from expert-specific notions of answer quality.}

% \begin{figure}[t]
%     \centering
%     \includegraphics[width=\linewidth]{figures/disagreement_distribution_pie_chart.png}
%     \caption{don't know if this should be included yet: Disagreement distribution}
%     \label{fig:disagreemnts}
% \end{figure}

% \begin{figure}[t]
%     \centering
%     \includegraphics[width=\linewidth]{figures/agreement_disagreement_by_category.png}
%     \caption{(don't know if this should be included yet) Expert agreement (green) and disagreement (red) by question type.}
%     \label{fig:disagreemnts2}
% \end{figure}

% \usepackage{booktabs}
\begin{table*}[]
\setlength{\tabcolsep}{6.5pt}
\centering
\scriptsize
\begin{tabular}{@{}lrr|rr|rr|rr|rr|rr@{}}
\toprule
                           & \multicolumn{2}{c}{\textbf{Gemini-flash}} & \multicolumn{2}{c}{\textbf{Gemini-pro}} & \multicolumn{2}{c}{\textbf{Claude-sonnet}} & \multicolumn{2}{c}{\textbf{Claude-sonnet-r}} & \multicolumn{2}{c}{\textbf{GPT-5.1}} & \multicolumn{2}{c}{\textbf{GPT-5} (test only)} \\ 
                           & agree.       & $\tau$-b        & agree.         & $\tau$-b           & agree.              & $\tau$-b               & agree.             & $\tau$-b              & agree.               & $\tau$-b               & agree.            & $\tau$-bs              \\\midrule
\rowcolor{lightgray} overall score            & 51.8\%       & 0.22       & 52.0\%         & 0.24          & 57.1\%              & 0.31              & 53.3\%             & 0.25             & 55.2\%               & 0.28              & 51.5\%            & 0.24             \\
answer relev.   & 29.6\%       & -0.07      & 30.4\%         & -0.12         & 24.3\%              & 0.00              & 25.8\%             & -0.14            & 40.5\%               & 0.01              & 39.2\%            & -0.03            \\
answer recall      & 45.9\%       & 0.15       & 48.1\%         & 0.16          & 47.3\%              & 0.14              & 48.4\%             & 0.15             & 48.7\%               & 0.17              & 44.6\%            & 0.10             \\
citation prec. & 40.5\%       & 0.20       & 41.0\%         & 0.19          & 42.5\%              & 0.25              & 40.8\%             & 0.19             & 38.9\%               & 0.20              & 43.3\%            & 0.23             \\
citation recall    & 40.2\%       & 0.15       & 42.9\%         & 0.21          & 44.4\%              & 0.25              & 43.4\%             & 0.18             & 43.8\%               & 0.26              & 42.3\%            & 0.22            \\ \midrule \midrule
                           & \textbf{all data}        & \textbf{agreed}   & \textbf{all data}        & \textbf{agreed}   & \textbf{all data}        & \textbf{agreed}   & \textbf{all data}        & \textbf{agreed}   & \textbf{all data}        & \textbf{agreed}   & \textbf{all data}        & \textbf{agreed}   \\
system corr ($\tau$-b)         & 0.40 & 0.60 & 0.40 & 0.69 & 0.48 & 0.70 & 0.40 & 0.60 & 0.40 & 0.73 & 0.53 & 0.73\\\bottomrule

\end{tabular}
\caption{Evaluation results using six alternative LLM judges: Gemini-2.5-flash (Gemini-flash, fresh run), Gemini-2.5-pro (Gemini-pro), Claude-sonnet-4.5 with (Claude-sonnet-r) and without reasoning (Claude-sonnet), GPT-5.1 without reasoning, and reasoning GPT-5. We limit reasoning GPT-5.1 run to test set due to cost constraints. We present the agreement (TOP) and the system correlations (BOTTOM) for both all data and expert-agreed subset.}
\label{tab:ablation}
\end{table*}

\section{Additional Findings}
\label{sec:additional-findings}

In this section, we briefly examine questions raised by our results in two additional findings.

\vspace{.5em}
\begin{mdframed}[backgroundcolor=gray!20] 
Finding 5: \textbf{Agreement statistics must be interpreted in the context of set of systems evaluated.}
\end{mdframed}

\noindent A side-by-side comparison of our IAA and overall preference agreement scores shows that our scores are generally lower than the values reported by our peers (Table~\ref{tab:related-works1}). This is in part explained by how ties are handled in our IAA and agreement calculations.\textsuperscript{\ref{footnote:iaa}}
We also find that the observed levels of IAA and model-human overall agreement is affected by how many systems are being evaluated and how similar they are in their quality. 

We compare IAA and model–human overall agreement across a subset of our evaluated systems varying in number and quality (analysis detailed in Appendix~\ref{sec:appendix-agreement-investigation}). 
When the evaluated systems are close in quality, preference judgments become more difficult, leading to lower agreement among human annotators and between evaluators and humans. 
Restricting the comparison to three more distinct systems (high-, medium-, and low-performing) yields substantially higher agreement. 
The results reported in Table~\ref{tab:score-agreements} are computed over six systems, including several of similar quality, which results in lower observed agreement.

\vspace{1em}
\begin{mdframed}[backgroundcolor=gray!20] 
Finding 6: \textbf{Our findings remain consistent across different LLM families \& evaluators.}
\end{mdframed}
\vspace{0.5em}

\noindent We also verify that our meta-evaluation results hold across a variety of different LLM evaluators for scoring of system outputs, beyond the one used by \sqaeval (gemini-2.5-flash). We rerun \sqaeval using six LLMs with large context windows, as required by the benchmark, over the same set of reports used in \S\ref{sec:meta-eval}. We then repeat the overall preference analysis (§\ref{sec:experimental_settings}). 

Across all major alternative LLM judges tested, overall agreements and correlations remain stable (Table~\ref{tab:ablation}). We observe variances at metric-level, most notably with the GPT models that see answer relevance score agreement (approx. 10-20 points) higher than other models. Otherwise, the results show agreements and correlation scores that are generally consistent with our earlier experiments. Moreover, computing the cross-correlations between the different model overall scores shows strong relationship  (Pearson \textit{r} of 0.73-0.92; Figure \ref{fig:llm-judge-cross-corr}). The results together indicate that our main findings are robust across evaluator choice, and not an artifact of a specific LLM judge.

\section{Recommendations}
Our findings reveal important limitations in current meta-evaluation practices and highlight the need for more nuanced approaches. We offer three key recommendations below.

\vspace{.2em} \noindent \textbf{Recommendation 1.} Human pairwise preference judgments is a tenable and effective approach at capturing relative performance of the system. However, our Findings 1 \& 2 suggest that comparison against \textit{preference} judgment should be limited to assessing at the system-level. For individual metric assessment, we recommend human annotation that is specifically designed to mirror the LLM evaluator instructions. 

% \vspace{.25em} \noindent \textbf{Recommendation 2.} Current deep-research report evaluation practices often center around the reporting system–level correlations or overall preference agreement with humans (Table~\ref{tab:related-works1}). Our findings surface important factors necessary for interpreting system-human agreement: metric-wise evaluation to disentangle annotator subjectivity from metric-based successes (Findings 2 \& 4), annotator expertise (Finding 3), and the set of systems evaluated, which influences the magnitude of agreement values (Finding 5). Further research may identify additional factors of importance. Alongside system-level or overall agreement scores, we recommend careful consideration of these contexts of evaluations. We also encourage reporting of disagreements, as greater transparency about these limitations acknowledges confounds in expert evaluation and helps identify avenues for improved meta-evaluation.

\vspace{.2em} \noindent \textbf{Recommendation 2.}  Our findings surface important factors necessary for interpreting system-human agreement: metric-wise evaluation to disentangle annotator subjectivity from metric-based successes (Findings 2 \& 4), annotator expertise (Finding 3), and the set of systems evaluated, which influences the magnitude of agreement values (Finding 5). %Further research may identify additional factors of importance. 
Alongside system-level or overall agreement scores, we recommend careful consideration of these contexts of evaluations. We also encourage reporting of disagreements, as greater transparency about these limitations acknowledges confounds in expert evaluation and helps identify avenues for improved meta-evaluation.

\vspace{.2em} \noindent \textbf{Recommendation 3.}  In line with our recommendation 2, we also emphasize the need to carefully match between annotator expertise and the evaluation goal. 
If the aim is to assess metrics that inherently require deep expertise, the study should recruit true domain experts who understand the topic's nuances and can accurately judge the detailed content of the report. In such cases, deep-expertise can be ensured by asking experts to generate their own questions, following clear guidelines to ensure the questions are substantive and non-trivial. Conversely, if the objective is to validate LLM-judges that simulate general user assessments, near-expert annotators may provide more reliable ground truth than deep-experts, whose specialized knowledge can introduce variability that diverges from the target population's judgments.

%Current practice in the research community leans heavily on reporting system-human agreements in the assessment of complex outputs like deep research reports. Our finding 4 suggest this approach may obscure the inherent challenges of subjectivity that permeates human evaluation of complex outputs. Our recommendation is that in addition to system-level scores, researchers should become more open about reporting disagreements as well. Greater transparency about these limitations would not only acknowledge the inherent confounders in expert judgment but also help identify practical steps for better meta-evaluation.

% \paragraph{Recommendation 4.} Finally, all LLM-based evaluations are susceptible to systematic biases introduced by the underlying model. Consequently, whenever possible, we recommend validating results across multiple evaluator families. \jena{don't know if we should include...}
% \dh{Seems like something people already know? and not sure it's one of the main takeaways from the work done in this paper, right?} \vk{can we check if other papers report >1 evaluators? I don't think so. So it might be worth saying?} \jena{They only do it in the context of comparing their own models and against baselines. I am inclined not to include this recommendation because it seems like it only applies for those instances where the work is making claims about the meta eval findings (which we do). For some papers only testing with one model may be sufficient (considering cost of eval as well and access to these families of models)?}

\section{Conclusion} 

%This work examined the validity of using human preference judgments as a meta-evaluation signal for LLM-based evaluators of complex research reports, using \sqaeval as our test case. Our analysis reveals several important findings. We find that overall preference rankings are best suited for system-level evaluation; explicit metric-wise annotation is necessary for assessing effectiveness of individual metrics in the evaluation;  annotator expertise significantly impacts evaluations; and human judgments exhibit substantial subjectivity, underscoring the inherent difficulty of the task. Our findings recommend targeted evaluation designs, careful selection of annotator expertise, and transparency around disagreement when validating automated evaluators for open-ended tasks.

We examine the validity of using human preference judgments as a meta-evaluation signal for LLM-based evaluators using \sqaeval. Results show that overall rankings are best suited to support system-level evaluation, metric-wise annotations are required for metric analysis, and annotator expertise significantly impacts evaluation outcomes. Our results also highlight the challenge of subjectivity in human judgments.  We recommend targeted evaluation designs, careful selection of annotator expertise, and transparency around disagreement when validating automated evaluators for open-ended tasks.

%Beyond meta-evaluation, our findings also hold implications for the future of deep-research evaluation: we propose that future evaluation lies in making the evaluations user context sensitive. Given that expert judgments reflect fundamentally different notions of quality rather than random noise or a common baseline expectation, standardized evaluation may be inherently limited in what it can capture. This points to the need for future evaluation frameworks that explicitly model and account for user expectations.

Beyond meta-evaluation, our findings suggest deep-research evaluation should be user-context sensitive. Inconsistencies across expert judgments often reflect their distinct preferences and notions of quality---rather than random noise.  Our findings indicate that a one-size-fits-all evaluation is inherently limited in what it can capture. This points to the need for future evaluation frameworks that explicitly model the diversity of user expectations.

\section*{Limitations}

Our meta-evaluation relies on relatively small expert-labeled datasets in both near- and deep-expertise settings, reflecting the high cost of recruiting qualified scientific annotators. In addition, the human evaluation is based on a limited pool of domain experts (five annotators), which may restrict the generalizability of our findings. Finally, like all LLM-based evaluation frameworks, our analysis may still be influenced by biases inherent to the underlying evaluator models.

% \section*{Reproducibility Statement}

% We took special care to make this work reproducible. We (plan to) open source code, annotated data, and other materials for reported meta-evaluation experiments to support reproducibility.

%\section*{Acknowledgments}

\bibliography{bibliography}

\appendix

\pagebreak
\section{Appendix: Experimental Details}
\label{app:expt_details}
%Pre-processing decisions, model parameters, feature templates, lengthy proofs or derivations, pseudocode, sample system inputs/outputs (including prompts), annotator guidelines, URLs  and other details that are necessary for the exact replication of the work described in the paper.

% \subsection{\sqaeval Details}
% \label{sec:appendix-sqa-eval-details}
% We anonymize the citation of the publication that details \sqaeval \cite{bragg2025astabenchrigorousbenchmarkingai} for the double-blind review process. In order to make the relevant information available to the reviewers, however, we have put together excerpts from the larger paper that is specifically relevant to \sqaeval in \href{https://www.dropbox.com/scl/fi/yvt8p9ozilfd6bdrygp7b/Content_of_Anonymous__2025.pdf?rlkey=mhy9l9y1kctjjs4qm0amy46x9&st=czh7z2wn&dl=0}{this anonymized document}. Please find a more in-depth discussion of the evaluation and all its supporting information (e.g. prompts and input/output samples) in this document.

\subsection{Annotation Procedure, Guidelines \& Interfaces}
\label{sec:appendix-guidelines}

\begin{figure*}[th!]
    \centering
    \fbox{\includegraphics[width=\linewidth]{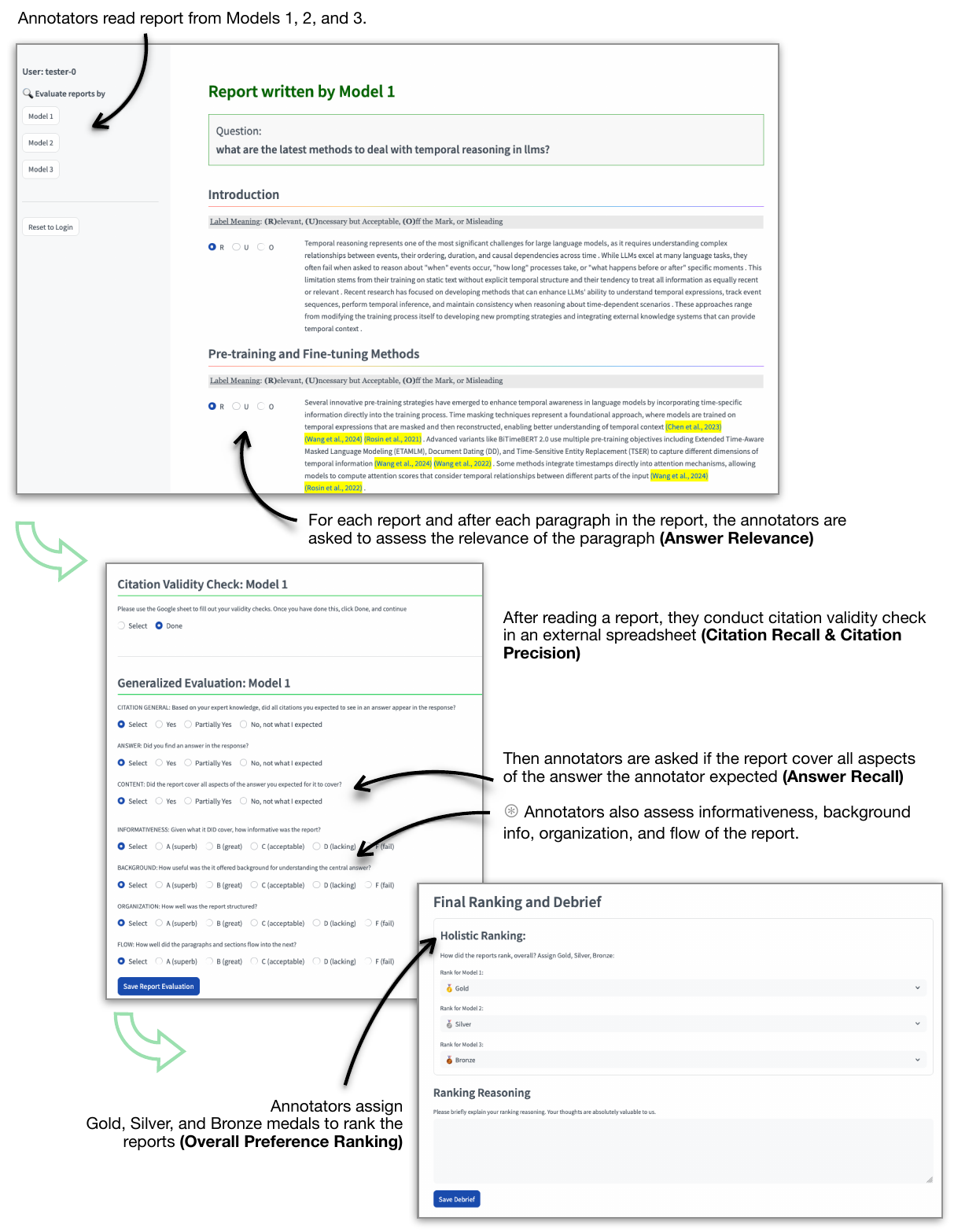}}
    \caption{Annotation Procedure and Screenshots}
    \label{fig:annotation-procedure}
\end{figure*}

The annotation procedure for 3-way evaluation that include overall preference ranking and metric-wise evaluation is illustrated in Figure~\ref{fig:annotation-procedure}. 

The complete set of instructions given to annotators this process is found in \href{https://github.com/allenai/ai2-scholarqa-eval/meta-evaluation/}{the project repository}. All instructions and human annotation results will be released with the code.

\subsection{Score Conversion \& Pairwise Calculation}
\label{sec:appendix-score-conversion}

\begin{figure}[t!]
\centering
  \includegraphics[width=.8\linewidth]{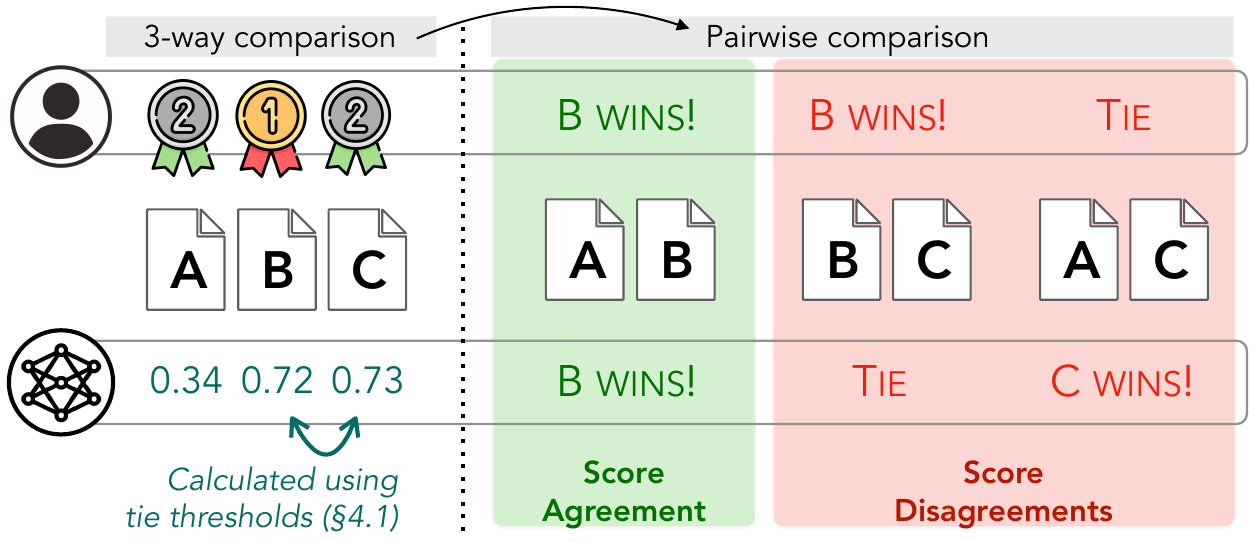}
  \caption{Human annotation is conducted as a 3-way comparison. The results are converted into pairwise judgments, and is compared to the model scores for score (dis)agreements.}
  \label{fig:score-agreement-calc}
\end{figure}

As illustrated in Figure~\ref{fig:score-agreement-calc}, we convert expert 3-way rankings into pairwise winner to enable human-model agreement analysis. In the 3-way setting, an expert provides a full ordering over reports A, B, and C, optionally allowing ties based on score thresholds. We use transitivity to decompose each 3-way ranking into the three implied pairwise comparisons, i.e., (A, B), (B, C), and (A, C), where we can declare pairwise winner (or tie) for each comparison. For model scores, the scalar scores are used to decide the pairwise winner and ties are computed using the tie thresholds (see next section). Finally, agreements are calculated by comparing the two scores.

\subsection{Tie Threshold Calibration}
\label{sec:appendix-thresholding}

Because human judgments allow ties, we introduce a corresponding tie decision for model outputs. Model scores are continuous, so we define a tie when the absolute difference between the two scores in a pair falls below a threshold. We select this threshold by matching the overall tie rate in the human data: for each set of data being compared, we choose the threshold that yields a proportion of model ties closest to the observed proportion of human ties. If multiple thresholds achieve the same match, we select the smaller threshold value.

\subsection{Answer Rubric Comparison to Baselines}
\label{sec:appendix-rubric-validation}

We formulate a task that compares our rubrics against baselines to determine how well our automated rubric creation approach (Section \ref{sec:sqaeval}) holds up against rubrics creation approaches. In particular, we are interested in investigate the comprehensiveness and utility of our content rubrics. Our rubric creation uses a report-pooling approach to extracting and merging ingredients into rubric approach. We compare it against the more common practices of rubric generation: human-written rubrics and generation of rubrics based on LLM's parametric knowledge.

\paragraph{Set up.} We source questions from the dev set, for which OpenScholar \citep{asai2024openscholar} provides question-specific, human-developed rubrics, which we use as one of our two baselines. Our second baseline comes from LLM-generated rubrics obtained using methods proposed by DeepResearch Bench framework \citep{du2025deepresearch}.

We follow the procedure described in the near-expert setting (Section \ref{sec:experimental_settings}) to have our experts select questions and conduct the preference ranking task. At the end of this set up, we introduce an additional step that shows the annotators the three rubrics associated with the question they are evaluating, and asks them to rank the rubrics on their quality. The quality is defined by by (1) coverage of important content, (2) comprehensiveness of the coverage, and (3) usability of the rubrics based on the judgment task they just completed. All five experts participated in this study with 5 selected questions each, leading each rubric participating in 50 matches.

\begin{figure}[t]
    \centering
    \includegraphics[width=\linewidth]{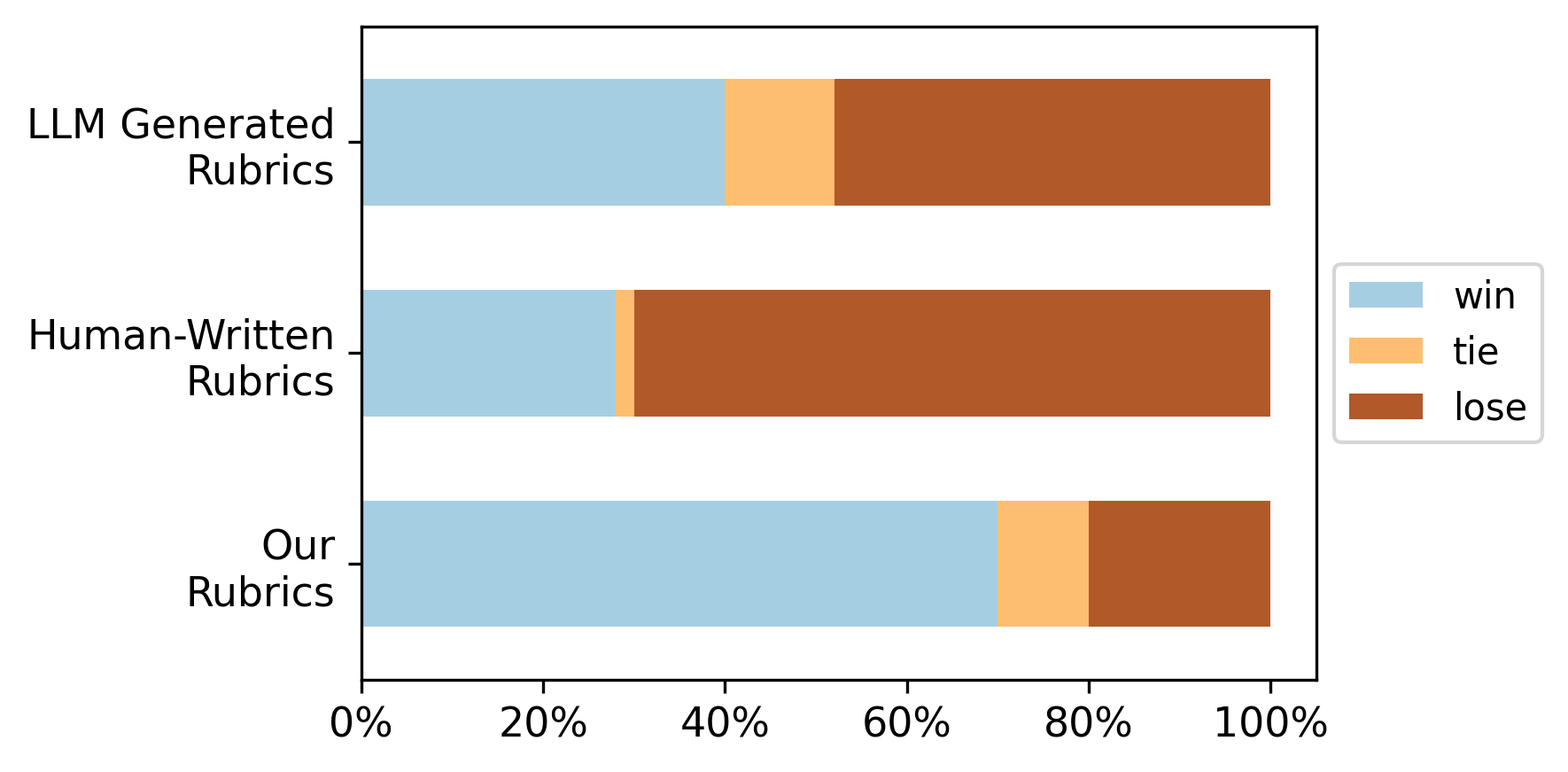}
    \caption{rubric comparison}
    \label{fig:rubric-comparison}
\end{figure}

\paragraph{Results.}
The pairwise match results show that our rubric wins 70\% of all its matches with 10\% ties. The win rate is significantly higher than the LLM rubrics (40\%; \textit{p=0.000}) and human-written rubrics (28\%; \textit{p=0.001}). 

\subsection{Citation Followup Study}
\label{sec:appendix-citation-followup-annot}

In our metric-wise experiment, we asked experts to select their own claims from the report, specifically focusing on the content most central to the answer to the question. Despite our careful efforts to instruct our experts on the claim selection to provide us with a comparable context, however, we found that this approach naturally leads to mismatch in claim selection relative to the LLM judge. Thus, we conduct a second set of citation annotations, where we control for selected claim span with the goal of enabling a more direct and methodologically consistent comparison.

\paragraph{Set up.} For this annotation, we recruit 15 researchers and scientists with expertise in NLP and AI. These experts were presented with a randomly assigned claim from a generated report given a test set question and their associated citations with excerpts. The researchers were instructed to mark a claim as \texttt{fully\_supported} only if every aspect of the claim was entirely supported by the cited sources, and to assess whether each citation in fact supported its associated claim. A skip option was made available for each instance of assessment, in order to provide opportunity to skip content beyond their expertise, but they were specifically instructed to avoid skipping when possible, so as not to bias the annotation to simpler instances (e.g., simpler claims, shorter citation list).

\begin{figure}
    \centering
    \includegraphics[width=1\linewidth]{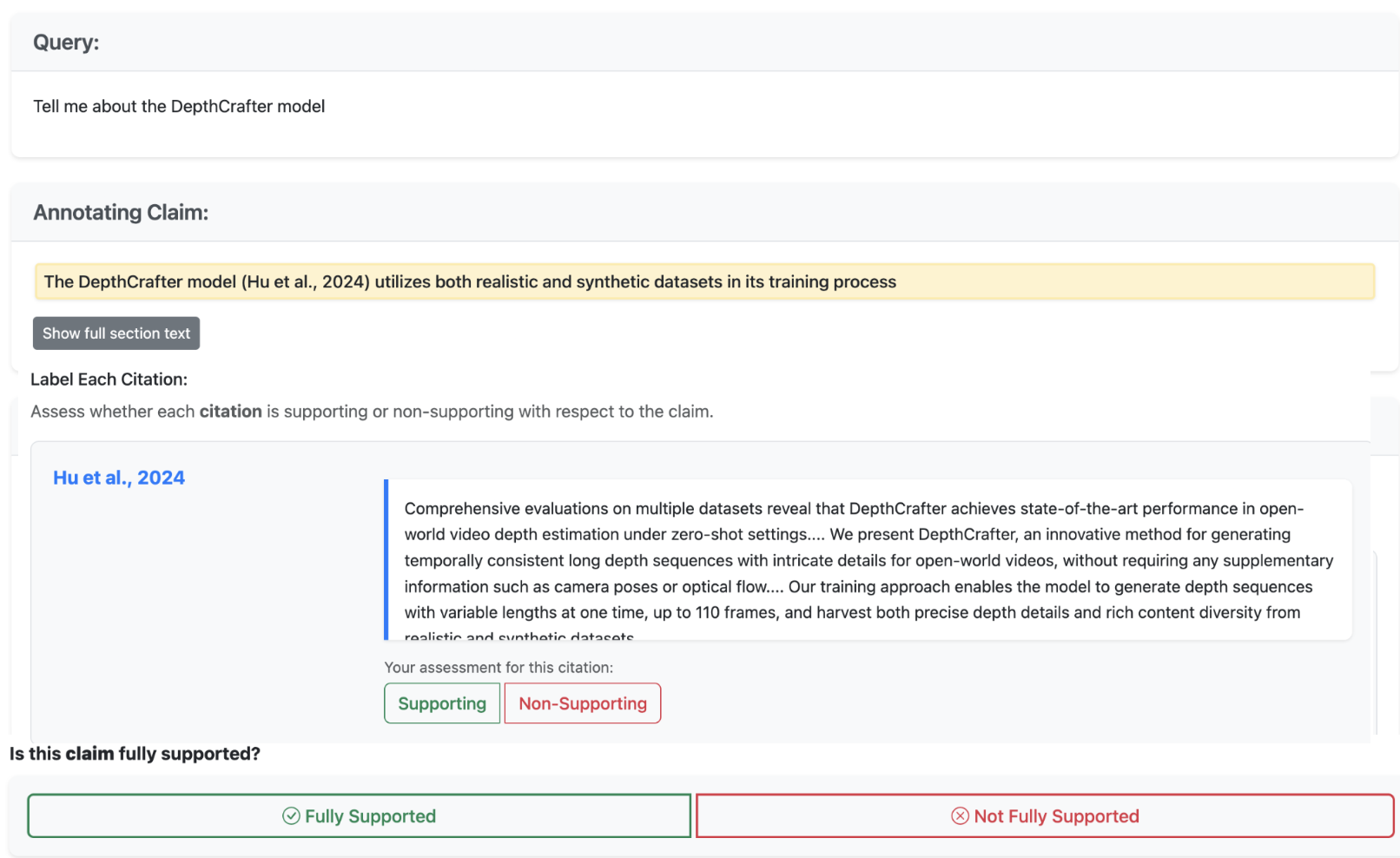}
    \caption{The annotation UI used to collect validation data from CS researchers regarding the quality of the citation evaluation quality.}
    \label{fig:eval-ui}
\end{figure}

Our 15 annotators assessed 80 claims along with their associated citations using this method. Figure \ref{fig:eval-ui} is a screenshot of the interface used for assessment.

\paragraph{Results.}
With stronger alignment between model evaluation set up to human one, we see an even stronger association between model scores and human. 
Overall, researcher assessments aligned with the LLM evaluator
69\% of the time on whether a claim was fully supported and 75\% of the
time on whether a citation supported a claim. In cases of
disagreement, the LLM was typically more lenient than human assessors:
85\% of discrepancies arose due to researchers labeling claims as not
fully supported while the LLM assessed the claim as fully
supported.

%Experts frequently reported confusion about claims produced by one particular system \aps{name and shame :P} included in our pool. After
Experts frequently reported confusion about claims produced by Elicit. After excluding claims from this system, agreement between the LLM and
researchers increased to 75\% on whether a claim was fully supported
and 81\% on whether a citation supported its associated claim,
indicating that much of the earlier disagreement was driven by unclear
or difficult-to-interpret claims rather than a fundamental misalignment
between human and automated evaluations.

\subsection{Relevance Adjudication}
\label{sec:appendix-relevance-adjudication}

To further investigate the role of expertise on human answer relevance, we design a small qualitative annotation experiment. 

\paragraph{Set up.} For this task, we select 10 questions from each setting (near- and deep-expertise settings) with the highest disagreement between human overall and model overall scores. We present the reports to the human evaluator with each disagreement highlighted. If model reasoning is available for the segment, we include it as well. For each highlighted disagreement, the expert is asked to review the text and adjudicate the disagreement by choosing to "Keep" their own answer, "Switch" to the model answer (i.e., they are persuaded by model's reasoning), or deem it a "Subjective" difference. Optionally, they are also asked to provide reasoning behind their choice. The 20 reports contain a total of 151 disagreements. 

\paragraph{Results.} The results of qualitative evaluation are summarized in Figure~\ref{fig:precision-disagreement-adjudicated}. Overall, the two settings (near- and deep-expertise settings) show proportionally similar levels of instances where the expert chooses to keep their own judgment. The difference lies in how willing experts are to adopt the model's answer choice or to attribute the disagreement as a matter of subjective judgment. Experts tend to defer to the model answer more often in the near-expert setting where the expertise-to-topic alignment is weaker, and model rationale serves to clarify (ir)relevance of fine-grained details. This behavior makes sense: in the near-expert setting, annotators do not necessarily have specific expertise in the task they are evaluating, so changing their minds based on further details is more expected than in the setting with experts have question-specific expertise. The differences between these two settings are statistically significant (\textit{t(149)=2.94, p=0.004}), and supports the earlier finding that for relevance judgment task is highly sensitive to evaluators' specific expertise.

%                         lose   win   tie
% rubric_name
% Human Written\nRubrics  0.20  0.70  0.10 
% Our\nRubrics            0.70  0.28  0.02 -11.35 1.4e-19
% LLM Generated\nRubrics  0.48  0.40  0.12 

%                        t(98) p-value
% ours to human written  -3.27 0.001
% ours to llm generated  -5.33 6.1e-07

% annotator based differences
% annotator tendencies

% \paragraph{Report Formatting} 
% We ensure consistent evaluation across different systems by standardizing all responses into a JSON format that decomposes each report into a list of sections paired with their accompanying citations. This transformation is either performed by the system submitting the response to be evaluated or using an LLM (gemini-2.5-flash) which takes the raw report text and maps it into our structured format. Further details are in §\ref{sec:appendix-prompts-n-formats}. \dd{+1 to Varsha's comment that we can cut this if we need space}

\subsection{Analysis on the Interaction between ~~ ~~ ~~ ~~ \# Systems in Evaluation vs. Agreement}
\label{sec:appendix-agreement-investigation}

\begin{table}[t!]
\footnotesize
\centering
\begin{tabular}{lllll}
\toprule
~ &~ & IAA & Agree. \\ \midrule
\multicolumn{2}{l}{2 systems} & 55.3\% & 43.4\%  \\
\multicolumn{2}{l}{3 systems} & 57.0\% & 44.6\%  \\ \midrule
\multicolumn{2}{l}{all 6 systems} & 55.0\% & 51.6\% \\
$\hookrightarrow$ & w/o Elicit & 57.0\% & 58.9\%  \\

\bottomrule
\end{tabular}
\caption{Comparing most competitive or "best" systems. \textit{\textbf{3 systems}}: \sqaeval, \sqaeval-SFT, OpenAI-DR; \textit{\textbf{2 systems}}: \sqaeval, \sqaeval-SFT.}
\label{tab:agreement-investigation-close}
\end{table}

\begin{table}[t!]
\footnotesize
\centering
\begin{tabular}{lllll}
\toprule
~ &~ & IAA & Agree. \\ \midrule
\multicolumn{2}{l}{2 systems} & 73.1\% & 83.7\%  \\
\multicolumn{2}{l}{3 systems} & 64.1\% & 71.4\%  \\ \midrule
\multicolumn{2}{l}{all 6 systems} & 55.0\% & 51.6\% \\
$\hookrightarrow$ & w/o Elicit & 57.0\% & 58.9\%  \\

\bottomrule
\end{tabular}
\caption{Comparing maximally, quality-wise, distinct systems. \textit{\textbf{3 systems}}: \sqaeval (high), OpenAI-DR (mid), Perplexity (low); \textit{\textbf{2 systems}}: \sqaeval (high), Perplexity (low).}
\label{tab:agreement-investigation-diff}
\end{table}

In this section, we investigate the impact of the quality and number of system included in human evaluation on IAA and human-model agreement. 

\paragraph{Set up.} For this study, we restrict the IAA and agreement comparison to 2 systems and 3 systems in two settings. Using final human evaluation ranking (Table~\ref{tab:final-rankings}) as a reference for quality, we select (1) the most competitive or "best" systems, and (2) the most distinct systems. The distinct systems are also represent heterogeneous systems in terms of the system goals and designs.

\paragraph{Results.} Restricting the comparison to best systems (Table~\ref{tab:agreement-investigation-close}) lowers the human-model agreement rate and the IAA does not see improvement. In fact, this set up leads to human-agreement rates that are consistently lower than the IAA. The drop more evident if we compare against values that exclude Elicit (the system that sees a systematic human-model disagreement). Restricting comparison to the most distinct systems, increases leads to a consistent improvement in both human-model agreement and IAA with a 2-system comparison leading to the highest agreement.

% \begin{table}[!ht]
% \footnotesize
% \centering
% \begin{tabular}{llllll}
% ~ & \multicolumn{2}{c}{Maximal Diff} & \multicolumn{2}{c}{Minimal Diff} \\ \cmidrule{2-3}\cmidrule{4-5}
% ~ & IAA & Agree. & IAA & Agree. \\ \midrule
% 2 systems & 73.1\% & 83.7\% & 55.3\% & 43.4\%  \\
% 3 systems & 64.1\% & 71.4\% & 57.0\% & 44.6\%  \\ \midrule
% 5 systems & 57.0\% & 58.9\% && \\ 
% \end{tabular}
% \end{table}

\pagebreak
\section{Appendix: Complementary Results}
\label{sec:appendix-compl-results}

\begin{table}[h!]
\centering
\footnotesize
\begin{tabular}{@{}lcc@{}}
\toprule
Evaluated Systems           & Expert & \sqaeval \\ \midrule
\sqaeval     & 1      & 1                       \\
\sqaeval\texttt{-SFT} & 2      & 3                       \\
OpenAI-DR    & 3      & 5                       \\
STORM        & 4      & 4                       \\
Elicit       & 5      & 2                       \\
Perplexity   & 6      & 6                      \\
\bottomrule
\end{tabular}
\caption{Final rankings based on human overall preference ratings for the full dataset.}
\label{tab:final-rankings}
\end{table}

\begin{figure}[h!]
\centering
    \includegraphics[width=.9\linewidth]{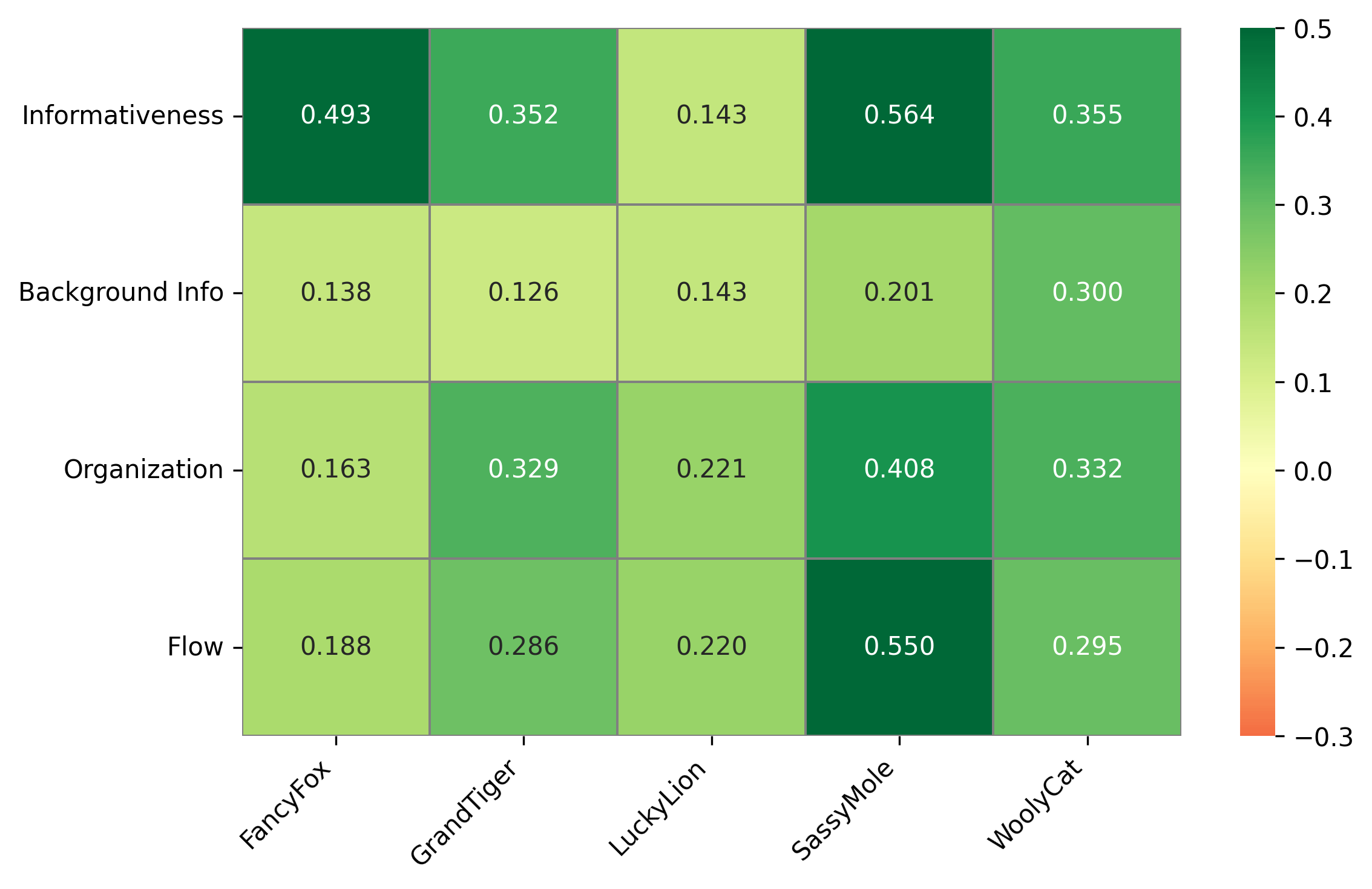}
    \caption{Analysis of qualitative assessments shows that each annotator calibrates what constitutes a good answer differently.}
    \label{fig:byannot-pref-vs-qualitative}
\end{figure}

\begin{figure}[h!]
\raggedleft
    \includegraphics[width=.9\linewidth]{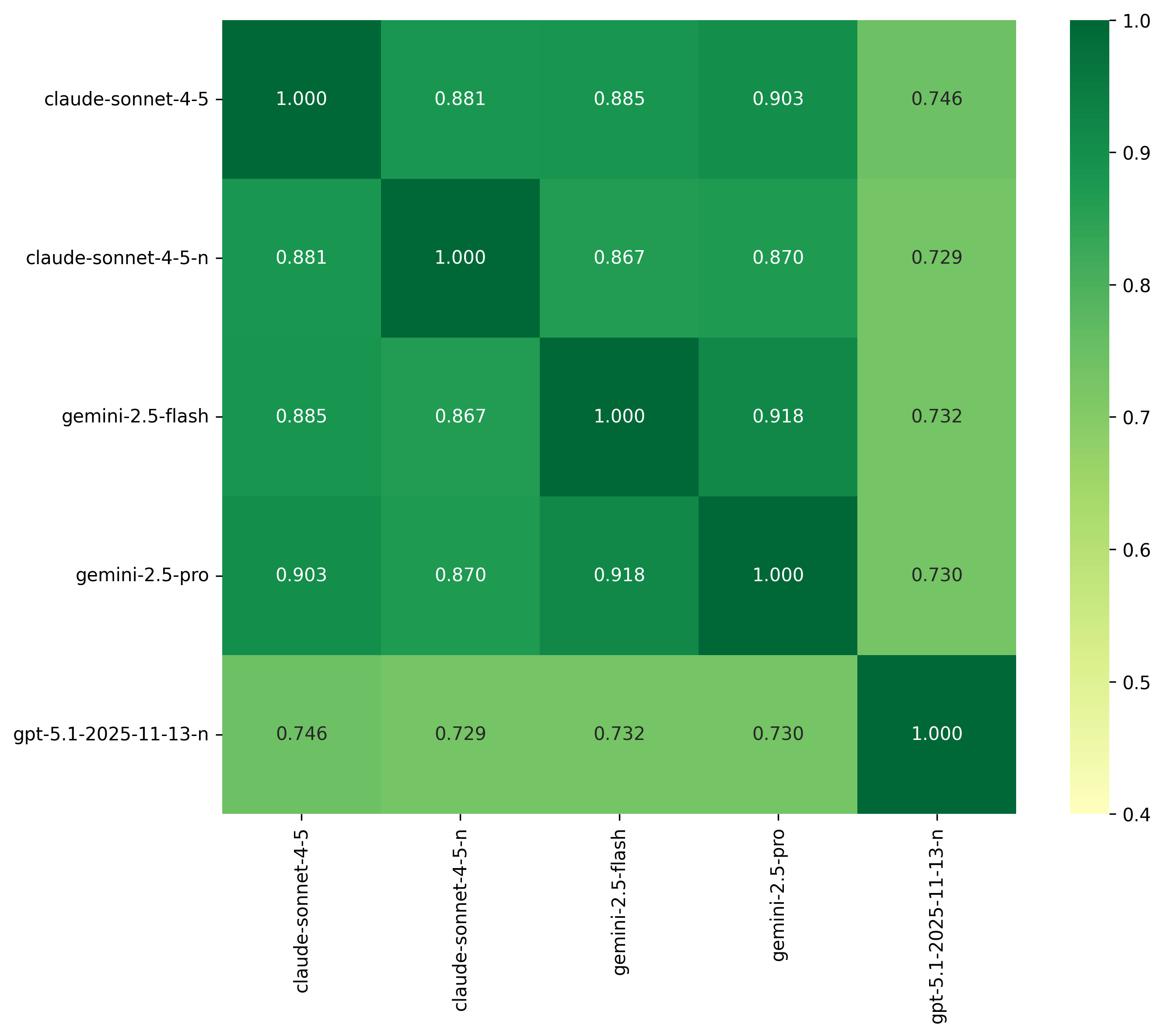}
    \caption{Cross-correlation heatmap for Pearson $\rho$ coefficient. Cross-correlation analysis indicates that all models are highly correlated with one another, and that our findings are robust across evaluator choice.  }
    \label{fig:llm-judge-cross-corr}

\end{figure}

\begin{table*}[ht]
\renewcommand{\arraystretch}{1.15}
\setlength{\tabcolsep}{3pt}
\footnotesize
%p{10mm}p{6mm}p{10mm}p{11mm}
\rowcolors{3}{}{lightgray!50}
\begin{tabularx}{\textwidth}{@{}p{29mm}lp{12mm}p{5mm}p{9mm}ccccc|ccc@{}}
\rowcolor{white}       &  &  & & &  \multicolumn{8}{c}{\textbf{Human Meta-Evaluation}}  \\ 
\rowcolor{white}       &  &  & & &  \multicolumn{8}{c}{\textbf{(Expert Agreement w/ LLM-judge)}}  \\ \cmidrule{6-13}
\rowcolor{white} 
\makecell[lt]{\\\textbf{Evaluation}}  &   
\makecell[lt]{\\\textbf{Dom}}      & 
\makecell[lt]{\textbf{Input}\\\textbf{Source}} & 
\textbf{Ext. Src} & 
\makecell[lt]{\textbf{Metrics}\\\textbf{Used}} & 
{\fontsize{8}{8} \makecell[lt]{\textbf{Expert}\\\textbf{Eval}}}      &  
{\fontsize{8}{8} \makecell[lt]{\\\textbf{PPR}}} & 
{\fontsize{8}{8} \makecell[lt]{\textbf{Metric}\\\textbf{-Wise}}} & 
{\fontsize{8}{8} \makecell[lt]{\textbf{Reported}\\\textbf{Comp.}}} & 
{\fontsize{8}{8} \makecell[lt]{\textbf{Qs}\\\textbf{Assign}}}   &
{\fontsize{8}{8} \makecell[lt]{\textbf{\# Sys}\\\textbf{Comp.}}} & 
{\fontsize{8}{8} \makecell[lt]{\\\textbf{IAA}}} &
{\fontsize{8}{8} \makecell[lt]{\\\textbf{OA}}}\\ \midrule

\rowcolor{yellow!20} \makecell[tl]{\sqaevalbl\\ \citep{bragg2025astabenchrigorousbenchmarkingai} \\ \ + present study}   & Sci:CS            & \makecell[tl]{real user\\queries}                     & \texttt{\small PI}                               & \texttt{\small AR,RCC, CP,CR }       &   $\checkmark$   &  $\checkmark$ &  $\checkmark$$\checkmark$ & \makecell[tc]{\texttt{~SC,IC,~~}\\\texttt{~OA,MA,~~}\\\texttt{MC}}& \texttt{\small R,C,A}  & 6 & 55.0\% & 51.6\%\\
\makecell[tl]{OpenScholar\\ \citep{asai2024openscholar}}                  & Sci          & written queries              & \texttt{\small PI}                                               & \texttt{\small RCC,CP, CR,AQ}        &  $\checkmark$  & $\checkmark$  & $\checkmark$* & \texttt{\small OA} & \texttt{\small R}  & 3 & 68\% & 50\%\\
\makecell[tl]{Deep Research Bench \\ \citep{Bosse2025deepresearchbench}}                  & Gen           & synthetic queries      & \texttt{\small WS}                                                 & \texttt{\small RC,CP}                  & - & - & - & - & - &&&\\
\makecell[tl]{DeepResearchGym\\\citep{coelho2025deepresearchgym}}               & Gen & \makecell[tl]{real user\\queries}  & \texttt{\small WI}                  & \texttt{\small FC,AQ, CP,CR }          & $\checkmark$ & $\checkmark$ & $\checkmark$* & \texttt{\small SC,OA} & \texttt{\small R}  & 7 & \makecell[tc]{0.87\\{\scriptsize Cohen $\kappa$}}  & \makecell[tc]{0.72\\{\scriptsize Cohen $\kappa$}}\\
\makecell[tl]{DeepResearch Bench\\\citep{du2025deepresearch}}                   & Gen             & curated queries       & \texttt{\small WS}                                                 & \texttt{\small RC,CP, CR}          &  $\checkmark$  & $\checkmark$ &  - & \texttt{\small SC,OA} & \texttt{\small R} & 4 & 68.4\% & 71.3\% \\
\makecell[tl]{\scriptsize DeepResearch-ReportEval\\\citep{Fan2025UnderstandingDV}} & Gen  &  curated queries & \texttt{\small GT, WS}      & \texttt{\small FC,AQ}           &  $\checkmark$  & $\checkmark$  & - & \texttt{\small OA} & \texttt{\small R} & 4 & - & 61.1\%\\
\makecell[tl]{LiveDRBench \\\citep{java2025livedrbench}}                  & Sci   & synthetic queries         & \texttt{\small WS}                                                &\texttt{\small  CP,CR }                 &  -      & - & - & - & - &&&\\
\makecell[tl]{ReportBench \\\citep{li2025reportbench}}                    & Gen           & synthetic queries    & \texttt{\small GT}                                             & \texttt{\small FC,CP, CR}              & -      & - & - & - & -&&&\\
\makecell[tl]{ResearchQA       \\\citep{yifei2025researchqa}}                 & Gen           & synthetic queries      & \texttt{\small WS}                                                  & \texttt{\small RCC,CR}                &   $\checkmark$ & $\checkmark$ &  $\checkmark$ & \texttt{\small OA,MA}  & \texttt{\small R}  & 3 & 84\% & 75\% \\
\makecell[tl]{\scriptsize DeepResearch Bench II\\\citep{Li2026DeepResearchBI}}                   & Gen             & \makecell[tl]{curated\\tasks}         & \texttt{\small WS}                                                 & \texttt{\small RC,CP, CR}          &  $\checkmark$  & -  &  $\checkmark$* & \texttt{\small OA} & \texttt{\small R}  &&&\\
\makecell[tl]{DeepScholar-Bench\\\citep{patel2025deepscholarbench}}               & Gen          & \makecell[tl]{extracted\\ abstracts}                & \texttt{\small GT, WS}                                     & \texttt{\small FC,CP, CR}              &   $\checkmark$ & - &  $\checkmark$* & \texttt{\small OA} & \texttt{\small R}  &&&\\
\makecell[tl]{Trec Rag 2024\\\citep{pradeep2024trecrag}}  & Gen                   & \makecell[tl]{written\\ topics}               & \texttt{\small GT}                                                     & \texttt{\small FC}                      &               $\checkmark$ & - & $\checkmark$ & \texttt{\small SC,MC} & \texttt{\small R}  &&&\\
\makecell[tl]{ResearchRubrics    \\\citep{sharma2025researchrubrics}}                 & Sci           & written queries              & \texttt{\small WS}                                                  & \texttt{\small RC}                      &   -      & - & - & - & - &&&\\
\makecell[tl]{DeepResearchEval   \\\citep{Wang2026DeepResearchEvalAA}}                  & Gen           & \makecell[tl]{synthetic \\tasks}     & \texttt{\small WS}                 & \texttt{\small FC,RC, CP}              & $\checkmark$ & - & $\checkmark$* & \texttt{OA} & \texttt{R} &&&\\
\makecell[tl]{LiveResearchBench   \\\citep{wang2025liveresearchbench}}                  & Gen           & written queries  & \texttt{\small WS}                 & \texttt{\small RC,CP, CR}              &      -      & - & - & - & - &&&\\
\makecell[tl]{ResearcherBench \\\citep{xu2025researcherbench}}                     & Sci:AI         & synthetic queries & \texttt{\small WS}                                                  & \texttt{\small RCC,CP, CR }           &   $\checkmark$     & - &  $\checkmark$* & \texttt{\small OA} & \texttt{\small R}  &&&\\
\makecell[tl]{FINDER\\\citep{zhang2025fargenuinelyusefuldeep}}               & Gen          & curated queries            & \texttt{\small WS}                                     & \texttt{\small RC,CP, CR}              &   $\checkmark$ & - &  $\checkmark$ & \texttt{\small MA,OA} & \texttt{\small R}  &&&\\
\midrule

\multicolumn{13}{l}{\tiny \makecell[tl]{
\textbf{COLUMNS: \hspace{2.3mm}} \textit{\textbf{Input Source:}} Sources of input (e.g., queries, topics) in the benchmark. "Curated" and "written" queries stand in for to "expert-curated" and "expert-written", respectively.\\ 
\hspace{14mm} \textit{\textbf{Expert Eval:}} Does the evaluation  use comparison to expert human annotation?;  \ \  \textit{\textbf{PPR:}} Pairwise preference ranking used for meta-evaluation; \\
\hspace{14mm} \textit{\textbf{Last 3 columns:}} For papers evaluating on pairwise-preference ranking (PPR) only, how many report-generating systems are compared (\textbf{\# Sys Comp.}), \\
\hspace{17mm} what is their reported inter-annotator agreement (\textbf{IAA}), and what is their reported human-model overall agreement (\textbf{OA})? \\
\textbf{ABBREVS: \hspace{3mm}} \textbf{\textit{External Sources:}} \{PI: Paper index, WI: Webpage index, WS: Web Search, GT: Groundtruth docs\}; \\ 
\hspace{14mm} \textbf{\textit{Metrics Used}:} \{FC: Fact Check,  RC: Rubric Coverage, RCC: Rubric Coverage restricted to content, CP: Citation Precision, CR: Citation Recall, AQ: Answer Quality\};\\ 
\hspace{14mm} \textit{\textbf{Metric-wise Meta Evaluation:}} human eval specifically designed for \{$\checkmark\checkmark$: all metrics\, $\checkmark$: a subset of metrics, $\checkmark$*: a subset of metrics, only overall score reported\};  \\
\hspace{14mm}  \textit{\textbf{Reported Comparisons:}} \{SC: system correlation, IC: instance correlation, OA: overall agreement, MA: metric-wise agreement, MC: metric-wise correlation\};\\
\hspace{14mm}  \textit{\textbf{Qs Assigned:}} \{R: randomly assigned Qs, C: expert chosen Qs, A: expert authored Qs\}\\
}}\\ \bottomrule
                                                 
\end{tabularx}
\caption{Related Deep Research Evaluation Literature. Ordered alphabetically by author.} %\aps{"Reading the description about PPR separate from the rest of the Meta eval description later is a bit jarring"} \jena{aman, could you say more about PPR thing?}}
\label{tab:related-works-complete}
\end{table*}

%\label{fig:appendix-llm-judge-cross-corr}

\end{document}